\documentclass[runningheads]{llncs}

\usepackage{eccv}

\usepackage{eccvabbrv}
\usepackage{algorithm}
\usepackage[noend]{algpseudocode} 
\usepackage{graphicx}
\usepackage{booktabs}
\usepackage{url}
\usepackage{grffile}
\usepackage{amsmath}
\usepackage{longtable}
\usepackage{multirow}
\usepackage{multicol}
\usepackage{tikz}
\usepackage{subcaption}
\usepackage{pgfplots, pgfplotstable}
\usetikzlibrary{patterns}
\usepackage{tabularx}
\usepackage{soul}
\usepackage{subcaption}
\usepackage{color, colortbl,xcolor}
\definecolor{Gray}{gray}{0.9}
\definecolor{lightpink}{rgb}{.95, 0.8, 0.8}
\definecolor{lightyellow}{HTML}{ecbd84}
\definecolor{lightblue}{rgb}{0.87, 0.94, 1}
\definecolor{lightgreen}{rgb}{0.88, 1, 0.88}
\definecolor{magentablue}{RGB}{138,43,226}

\newcommand{\verytiny}{\fontsize{4pt}{6pt}\selectfont}
\definecolor{ao(english)}{rgb}{0.0, 0.5, 0.0}

\usepackage[accsupp]{axessibility}  

\usepackage{hyperref}

\usepackage{orcidlink}

\begin{document}

\title{\texttt{X-InstructBLIP}: A Framework for Aligning Image, 3D, Audio, Video to LLMs and its Emergent Cross-modal Reasoning} 

\titlerunning{\texttt{X-InstructBLIP}}


\author{Artemis Panagopoulou\inst{1}\thanks{Work done while interning at Salesforce Research, $^{\ast\ast}$Equal mentorship contribution.} \and
Le Xue\inst{2,\ast\ast} \and
Ning Yu\inst{2, \ast\ast} \and Junnan Li\inst{2} \and Dongxu Li\inst{2} \and Shafiq Joty\inst{2} \and Ran Xu\inst{2} \and Silvio Savarese\inst{2}\and Caiming Xiong\inst{2} \and Juan Carlos Niebles\inst{2}}

\authorrunning{A.~Panagopoulou et al.}

\institute{University of Pennsylvania \\
 \email{artemisp@seas.upenn.edu}\and
Salesforce AI Research \\}

\maketitle

\begin{abstract}
Recent research has achieved significant advancements in visual reasoning tasks through learning image-to-language projections and leveraging the impressive reasoning abilities of Large Language Models (LLMs). This paper introduces an efficient and effective framework that integrates multiple modalities (images, 3D, audio and video) to a frozen LLM and demonstrates an emergent ability for cross-modal reasoning (2+ modality inputs). Our approach explores two distinct projection mechanisms: Q-Formers and Linear Projections (LPs). Through extensive experimentation across all four modalities on 16 benchmarks, we explore both methods and assess their adaptability in integrated and separate cross-modal reasoning. The Q-Former projection demonstrates superior performance in single modality scenarios and adaptability in joint versus discriminative reasoning involving two or more modalities. However, it exhibits lower generalization capabilities than linear projection in contexts where task-modality data are limited. To enable this framework, we devise a scalable pipeline that automatically generates high-quality, instruction-tuning datasets from readily available captioning data across different modalities, and contribute 24K QA data for audio and 250K QA data for 3D. To facilitate further research in cross-modal reasoning, we introduce the DisCRn (\textbf{Dis}criminative \textbf{C}ross-modal \textbf{R}easo\textbf{n}ing (DisCRn)) benchmark comprising 9K audio-video QA samples and 28K image-3D QA samples that require the model to reason discriminatively across disparate input modalities. Code and data is available at \href{https://github.com/salesforce/LAVIS/tree/main/projects/xinstructblip}{https://github.com/salesforce/LAVIS/tree/main/projects/xinstructblip}.
\end{abstract}

\section{Introduction}
\label{sec:intro}

Humans inherently process information from multiple sensory modalities to interpret their surroundings and make decisions based on a comprehensive view of their environment. However, Multimodal Large Language Models (MLLMs) are primarily concentrated on visual tasks, often overlooking the rich diversity of other common modalities like Audio, Video, and 3D, and failing to tap into the potential of comprehensively understanding multiple modalities (>2) in unison, which is crucial for advanced tasks such as cross-modal reasoning\footnote{~\textit{Cross-modal reasoning} is the ability to \underline{integrate} and \underline{discriminate} information from multiple modalities over text, in contrast to ``multimodal reasoning,” traditionally reserved for vision-language tasks.}.

The incorporation of various modalities beyond images into LLMs is still an area ripe for exploration, particularly regarding effective integration frameworks. A significant challenge lies on the lack of instruction-tuning datasets for other modalities like Audio, 3D, and Video, especially for data that involve two or more modalities simultaneously, making joint modality training a plausible but resource intensive approach to enable cross-modal reasoning.

\begin{figure}[tb]
    \centering
\includegraphics[width=0.9\textwidth]{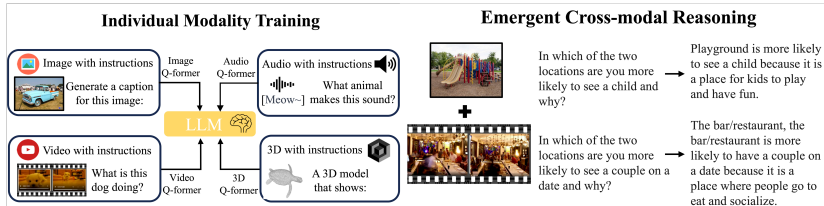}
    \caption{Despite utilizing distinct pre-trained encoders for each modality and independently aligning them to language through individual instruction aware Q-Formers, X-InstructBLIP demonstrates emergent abilities in cross-modal comprehension.}
    \label{fig:overview}
\end{figure}

In response to the above challenges, we introduce X-InstructBLIP,
  an extendable framework - illustrated in Figure \ref{fig:overview} and further analyzed in Section \ref{sec:method} - designed to align various modalities (image, 3D, audio, video) to LLMs, achieving single-modal reasoning tasks for each modality and enabling cross-modal reasoning across \textit{three or more modalities}. To facilitate this exploration and given the scarcity of unary instruction-tuning data for a spectrum of modalities other than the image modality, we introduce a simple yet potent approach in Section \ref{sec:fine-tuning_datasets}: a three-stage-query data augmentation technique to leverage \textit{open-source} LLMs to extract instruction-tuning data from captioning datasets.

\begin{figure}[tb]
    \centering    \includegraphics[width=\textwidth]{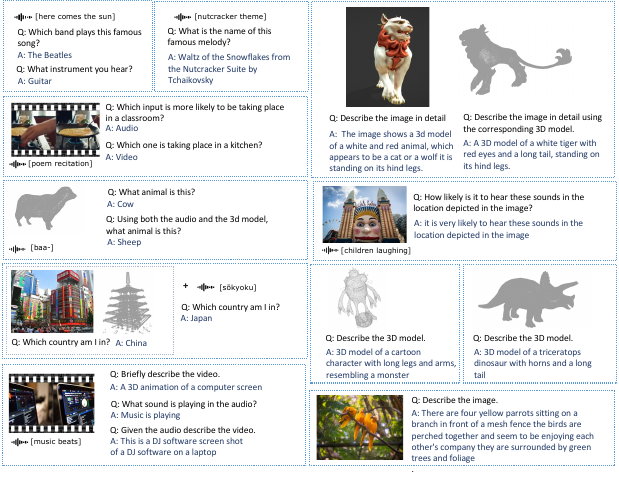}
    \caption{Qualitative Examples: X-InstructBLIP framework effectively handles both unimodal and cross-modal tasks without training on joint data.}
    \label{fig:qualitative}
\end{figure}

Our framework explores two state-of-the-art projection mechanisms on frozen LLMs -  a prerequisite for maintaining separate modality training - instruction-aware Q-Formers\cite{dai2023instructblip} and linear projections\cite{liu2023visual}. Through an expansive evaluation on 13 benchmarks across 4 modalities we find that Q-Formers tend to exhibit higher performance on single modality tasks and versatility in distinguishing when to reason in a joint or discriminative manner in the presence of 2+ extra-linguistic modalities. Figure \ref{fig:qualitative} shows illustrative results, highlighting the capabilities of our framework. To quantify and challenge this emergent ability we introduce \texttt{DisCRn} in Section \ref{sec:discriminatory_reasoning_dataset}, an automatically curated \textbf{Dis}criminatory \textbf{C}ross-modal \textbf{R}easo\textbf{n}ing challenge dataset requiring models to distinguish between diverse combinations of modalities, such as audio-video and 3D-image.

\noindent Our contributions are summarized as follows:\\ 
(i)
We introduce an extendable framework that aligns Image, 3D, Audio, and Video to LLMs, and we benchmark its emergent cross-modal reasoning capability across two projection mechanisms. This framework does not need specific pre-training tailored to each modality. To the best of our knowledge, this is the first attempt to demonstrate that discriminative cross-modal reasoning emerges naturally through individual modality alignment to LLMs.

\noindent (ii)
We introduce an automatic approach for crafting instruction-tuning datasets for a variety of modalities, leveraging only readily available captioning data and open-source language models. Contributing $\sim$ 250k samples for 3D QA data and $\sim$ 24k samples for Audio QA data.\\
\noindent (iii)
We collect \texttt{DisCRn}, the first dataset designed for evaluating instruction-based cross-modal discriminative reasoning. Which includes $\sim$36k examples across various modalities such as video, audio, 3D, and images.

\section{Related Work}
\textbf{Vision Language Models:} Recent years have seen a surge in models capable of executing a spectrum of vision-language tasks, leading to the creation of Multimodal Language Models (MLMs). These models align the static vision and language representations through various techniques, such as unified pre-training~\cite{cho2021unifying, wang2022ofa, yang2022unitab, li-etal-2022-mplug,wang2022git,Wang_2023_CVPR,huang2023language,sun2024emu,wang2023image}, vision-to-language alignment through textual feature extraction~\cite{yang2022empirical,gui2022kat,lin2022revive,shao2023prompting,wang2023filling}, vision-encoder optimization~\cite{tsimpoukelli2021multimodal},  and linear~\cite{liu2023visual,koh2023grounding,pmlr-v202-driess23a}, transformer-based~\cite{najdenkoska2023meta,mañas2023mapl,chen2022visualgpt}, or auto-encoder based projections~\cite{liu2023language,yu2023spae}. More relevant to this work are approaches that learn intermediate vision-informed language token representations either interleaved in LLM layers such as in Flamingo\cite{alayrac2022flamingo} and LLaMA adapter\cite{zhang2024llamaadapter} or only to the input layer such as in the BLIP series\cite{li2023blip,pmlr-v162-li22n,dai2023instructblip} which employ Q-Former based projections, LLAVA\cite{liu2023visual}, and MiniGPT4\cite{zhu2024minigpt} which employ linear projections. 

\noindent\textbf{Cross-Modal Language Models:} Projection-based approaches, initially focused on images, have recently broadened to encompass audio\cite{kim2023prefix,deshmukh2023pengi,tang2024salmonn,gong2024listen}, video\cite{yang2022zero,bain2021frozen,maaz2023videochatgpt,luo2024valley}, and 3D projections\cite{hong2023dllm,xu2023pointllm,guo2023point} into pre-trained large language models (LLMs). This expansion has seen the advent of unified pretraining frameworks such as mPLUG2\cite{xu2023mplug2} and OnePeace\cite{wang2023one}, as well as projection-based methods for enhancing frozen LLMs, like VideoLLaMA\cite{zhang2023video} and X-LLM\cite{chen2023x}, which aim to jointly train audio and video processors. Notably, X-LLM focuses on this integration primarily during the latter stages of training. In a similar vein, ChatBridge\cite{zhao2023chatbridge} adopts a training approach akin to X-LLM but utilizes a perceiver-based projection\cite{jaegle2021perceiver}. Audio-Visual LLM\cite{shu2023audio} follows a similar training paradigm of a final joint finetuning stage, but instead of maintaining a frozen LLM, it updates it using LoRA\cite{hu2021lora}. Our method is set appart by maintaining independent finetuning throughout and a frozen LLM avoiding the instability in training due to disparately aligned modality projections. Another line of work, including ImageBind-LLM\cite{han2023imagebind}, PandaGPT\cite{su-etal-2023-pandagpt} and PointLLM\cite{guo2023point} leverages unified representations such as ImageBind\cite{Girdhar_2023_CVPR} to only implicitly align additional modalities to LLMs by only training on image-text pairs.  Contemporary works such as AnyMAL\cite{moon2023anymal} 
 and OneLLM\cite{han2023onellm} have pushed the boundaries further by extending the application of projection-based approaches to additional modalities, such as 3D. Unlike other models that keep the LLM frozen, both opt to unfreeze the LLM during training. OneLLM adopts a router-based mixture of experts strategy to learn the mapping between different modalities. In contrast, AnyMAL focuses on jointly learning a LLaVA-style Projection layer for each modality during a portion of the training process. 

\noindent\textbf{Multimodal Multi-Input Language Tasks:} The advancements in single input vision-language tasks have paved the way for the development of tasks necessitating models to concurrently reason about multiple non-linguistic inputs, such as engaging in spatial reasoning across multiple images~\cite{bansal2020visual}, deliberating over a series of slides~\cite{SlideVQA2023}, responding to queries necessitating cross-modal reasoning across images and tables~\cite{li-etal-2022-mmcoqa}, or executing a range of instruction-based tasks involving multiple image inputs~\cite{li-etal-2022-mmcoqa}. Despite their complexity, these tasks operate mostly within the realms of image-text modalities. Even though cross-modal tasks exist, predominantly requiring models to reason over joint audio and video~\cite{alamri2018audio,li2022learning},
 there is a gap in the evaluation of models' generative capabilities in reasoning about cross-modal inputs \textit{contrastively}. While models are often optimized on contrastive objectives~\cite{chen2020simple,chen2020simclr,li2023blip,he2020momentum,jiang-etal-2023-vision,radford2021learning,li2020oscar,li2021align}, even in cross-modal settings~\cite{Girdhar_2023_CVPR,guzhov2022audioclip,nagrani2022learning}, their evaluation is confined to classification tasks or utilizing the contrastively learned representations for downstream tasks. To address this gap, we introduce \texttt{DisCRn}, a task requiring contrastive reasoning across cross-modal inputs in an open generation setting, evaluating a model's ability to translate features of various modalities from its internal representations to its generative output distribution.

\section{Method}
\label{sec:method}
\noindent \textbf{Framework Overview:} Figure \ref{fig:overview} depicts an overview of the framework's setup which extends instruction finetuning for image alignment\cite{dai2023instructblip, liu2023visual} to an arbitrary number of modalities through independent fine-tuning of modality-specific projections to a frozen LLM, further broken down in Algorithm \ref{alg:method}.
X-InstructBLIP's alignment framework involves the following steps: \textbf{(1)} For each modality, collect an instruction tuning dataset suite $(x,y) \in \mathbb{D}_M$ s.t. $x = (x_M, x_T)$ is a tuple of a modality input and text, and $y$ is the expected text output. \textbf{(2)} Let $\text{Enc}_M$ be a modality encoder to $R^{d_M}$ and $\text{Enc}_T$ be a mapping from text to the LLM's embedding space $R^{d_L}$. Optimize a single separate projection module $f^M_\theta: R^{d_M} \rightarrow R^{kd_L}$ for each modality $M$ on $\mathbb{D}_M$ while maintaining the parameters of the LLM frozen, 
where $k$ is the number of LLM input tokens corresponding to the non-linguistic input. For sequential data, such as video and audio, we extract $N\times k$ query tokens; each frame is encoded and processed separately by the projection module.  \textbf{(3)} \noindent The model is optimized under a causal language modeling  objective~\cite{j.2018generating}: 
{\footnotesize $\min_{\theta} \mathcal{L_{\text{CE}}}\Bigl(\text{LLM}(x_{\text{LLM}}), \text{h}(y)\Bigr)$}
where $\mathcal{L_\text{CE}}$ is the cross entropy loss, $\theta$ the Q-Former parameters, $y$ is the target sequence, $\text{LLM}(x_{\text{LLM}})$ is the LLM's prediction. 
\begin{algorithm}
\footnotesize
\caption{X-InstructBLIP Optimization Framework}
\label{alg:method}
\begin{algorithmic}[1]
\Require Set of modalities $\mathbb{M}$, each associated with a set of datasets $\mathbb{D}_M$, and set of templates $\mathbb{I} = \{I_{M_t}: M\in \mathbb{M}, t\in \mathbb{T}\}$ for each task $t\in \mathbb{T}$

\For{each modality $M$ in $\mathbb{M}$}
    \State Initialize modality-specific pre-trained encoder $\text{Enc}_{M}$
    \State Initialize LLM encoder $\text{Enc}_{T}$ (tokenize and embed text)
    \State Initialize projection $f^{M}_\theta: R^{d_M} \rightarrow R^{kd_L}$
    
    \For{each step in number of iterations}
        \State Sample $(x,y)$ from $\cup \mathbb{D}_M$
        \State Sample $i_{M}$ from $I_{M_t}$ where $t$ is the task mapping $x$ to $y$
        \State $z_{M} \gets \text{Enc}_{M}(x)$ \Comment{Encode input to embedding space}
        \State $w_{M} \gets f^M_\theta(z_M)$ \Comment{Transform encoded input to LLM embedding space}
        \State $x_{\text{LLM}} \gets w_{M} \Vert \text{Enc}_T(i_M) \Vert \text{Enc}_T(x_T)$
        
        \State $\text{Prediction} \gets \text{LLM}(x_{\text{LLM}})$ \Comment{Get LLM's prediction}
        \State $\text{Loss} \gets \mathcal{L_{\text{CE}}}(\text{Prediction}, \text{Enc}_T(y))$ \Comment{Calculate cross-entropy loss}
        \State $\theta \gets \theta - \alpha \nabla_\theta \text{Loss}$ \Comment{Update projection parameters}
    \EndFor
\EndFor

\end{algorithmic}
\end{algorithm}

 \begin{figure}[tb]
\centering
\begin{subfigure}[a]{.6\textwidth}\centering
   \includegraphics[width=\textwidth]{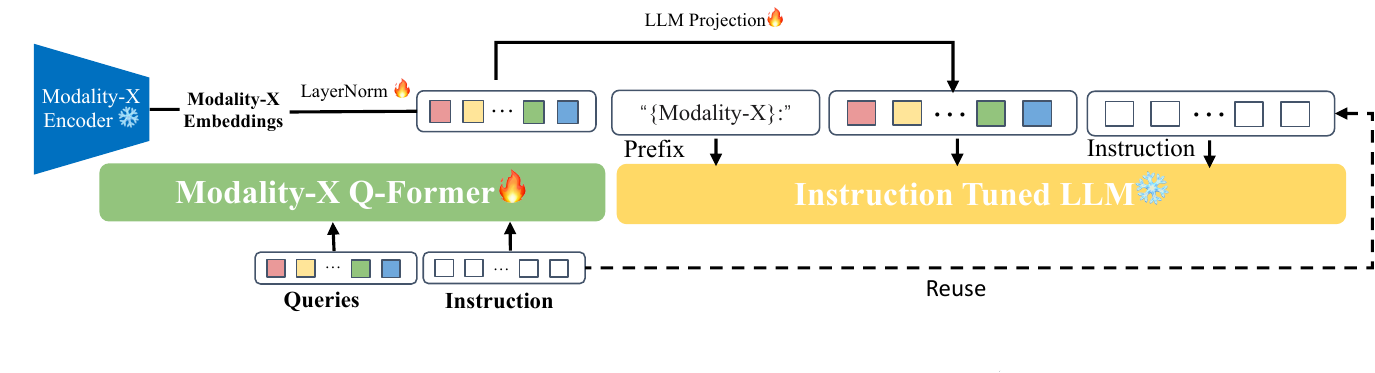}
   \caption{\textbf{X-Instruct Projection}}
   \label{fig:Q-Former_v_linear_proj}
\end{subfigure}
\begin{subfigure}[b]{.5\textwidth}\centering
   \includegraphics[width=\textwidth]{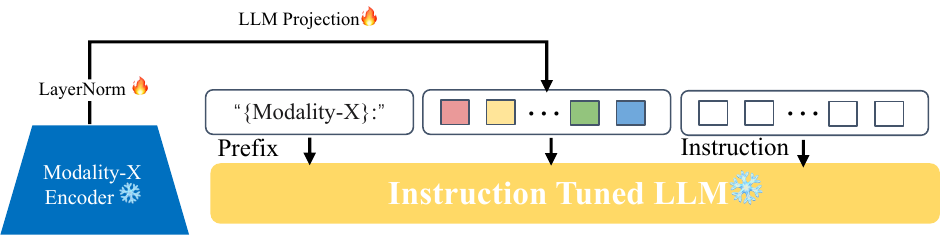}
   \caption{\textbf{X-LLaVA-style Projection}}
   \label{fig:linear_proj}
\end{subfigure}
\caption[Projection Types]{Projection types explored in the X-InstructBLIP Framework. (a) is an an instruction aware Q-Former projection\cite{dai2023instructblip} and (b) is  a linear projection\cite{liu2023visual}.}
\end{figure}

\noindent \textbf{\colorbox{gray!10}{X-Instruct Projection:}} Figure \ref{fig:Q-Former_v_linear_proj} highlights all components associated with learning instruction aware Q-Former projections\cite{dai2023instructblip} for multiple modalities. Given a modality $M$ encoding $z_{M} = \text{Enc}_{M}(x_{M})$ and task instruction $i_M\in \mathbb{I}_{M_t}$, the Q-Former module transforms a set of $k$ learnable embeddings $\mathbf{Q}_M = \{\mathbf{q}_{M_1}\ldots \mathbf{q}_{M_K}\}$ termed \textit{input query tokens} into instruction-aware  representations of $\mathbf{Q}'_M = \text{QF}_M(\mathbf{Q}_M, z_{M},i_M)$. The Q-Former module consists of two transformer submodules that share the same self-attention layers: one submodule interacts with the output of the modality encoder $\text{Enc}_{M}$ and the other is a $\text{BERT}_{\text{base}}$ text transformer that serves as both an encoder and decoder. 
Each Q-Former is initialized with the pre-trained weights from BLIP-2~\cite{li2023blip}, without the cross-attention layers due to a dimension mismatch between the image encoder in BLIP-2 and the other modality encoders. The modality embedding $z_{M}$ interacts with the instruction text $i_M$ and input query tokens $\mathbf{Q}_M$ via cross-attention layers inserted every other transformer block, yielding the \textit{output query tokens} $\mathbf{Q}'_M$ which are linearly projected to the frozen LLM's space through a learnable projection layer $\text{LP}_M$ specific to each modality. 
Let $\text{pf}_M$ the modality prefix, $x$ the example text input, and $y$ the target phrase. With {\footnotesize$\mathbin\Vert$} denoting concatenation, the \textit{LLM input tokens}  are: {\footnotesize $ x_{\text{LLM}} = \text{Enc}_T(\text{pf}_M)
    \mathbin\Vert \text{LP}_M(\mathbf{Q}'_M) \mathbin\Vert \text{Enc}_T(i_M) \mathbin\Vert \text{Enc}_T(x_T))$}.

\noindent\textbf{\colorbox{yellow!10}{X-LLaVA-style Projection:}} We implement an adaptation of LLaVA's architecture\cite{liu2023visual} to cater multiple modalities, similarly to the instruction aware Q-Former. Figure \ref{fig:linear_proj} depicts the architecture of this simple projection which linearly transforms the outputs of the modality encoder directly to  the input embedding space of the LLM. Formally, the model consists of a single linear projection layer $\text{LP}_M: R^{d_M} \rightarrow R^{kd_{\text{LLM}}}$, where  $d_{\text{LLM}}$ is the LLM's embedding dimension. To compare the two projection types we match the number of trainable parameters for each modality, and maintain the 
training set-up.  

\section{Datasets}
\label{sec:data}
\begin{figure}[tb]
    \centering
    \includegraphics[width=\textwidth]{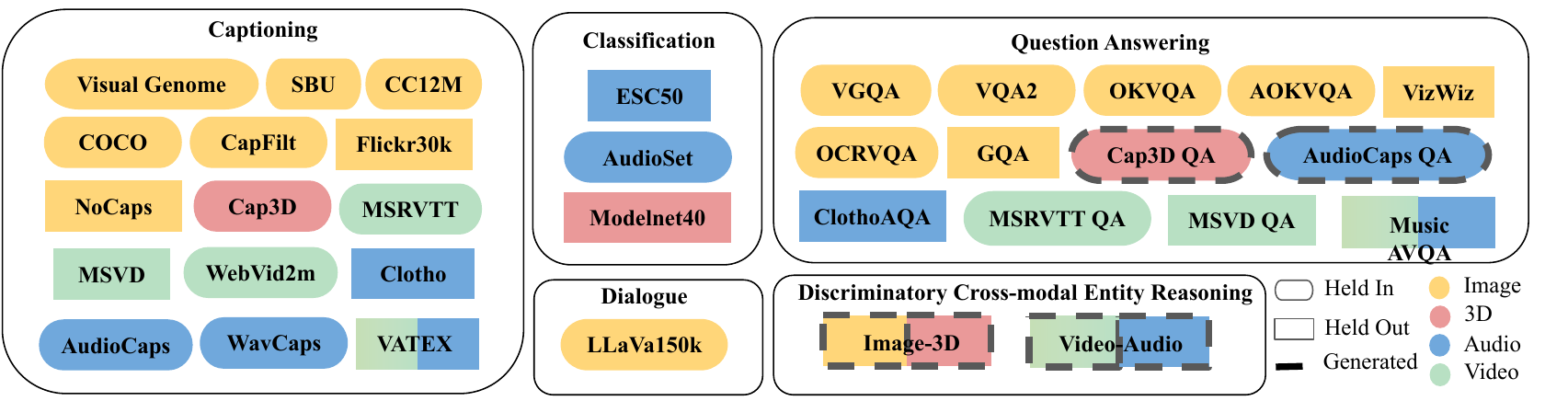}
    \caption{Instruction Tuning and Evaluation Datasets:  
    Oval-enclosed and square datasets are tuning and out-of-domain evaluation datasets respectively. Dashed outline is used for automatically derived datasets as described in Section \ref{sec:fine-tuning_datasets}.}
    \label{fig:instruction_datasets}
\end{figure}

X-InstructBLIP is optimized and evaluated on a collection of pre-existing and automaticaly generated datasets succintly presented in Figure \ref{fig:instruction_datasets}, discussed in Section \ref{sec:fine-tuning_datasets}, with more details available in the supplementary material. Section \ref{sec:discern_eval} introduces the Discriminatory Cross-modal Reasoning challenge dataset \texttt{\small DisCRn}\footnote{~The term \textit{discriminative reasoning}, adapted from~\cite{xu-etal-2021-discriminative}, refers to the ability to distinguish between sets of inputs, as opposed to \textit{joint reasoning}, the synthesis of information from aligned sources.} used to evaluate the emergent abilities of X-InstructBLIP~(Section \ref{sec:discriminatory_reasoning_dataset}).

\subsection{Fine-tuning Datasets}
\label{sec:fine-tuning_datasets}
\textbf{Existing Datasets:} Figure \ref{fig:instruction_datasets} illustrates the datasets utilized for both instruction tuning and evaluation. A detailed breakdown of the dataset statistics and formats can be found in the supplementary material. 
 For each dataset in \(\mathbb{D}_M\), the collection of held-in datasets specific to modality \(M\), a modified sampling strategy from~\cite{dai2023instructblip} is adopted accommodating a broader range of modalities. The sampling probability for any given dataset \(D_{M_d}\in \mathbb{D}_M\) is {\footnotesize\(\frac{\sqrt{|D_{M_d}|}}{\sum_{d\in [1\ldots |\mathbb{D}_M|]} \sqrt{|D_{M_d}|}}\)}, with minimal adjustments as justified in the supplementary matierial. 

\noindent\textbf{Instruction Data Augmentation:} Extracting instruction-aware representations necessitates diverse instruction-related tasks across all modalities. Notably, datasets for 3D and audio modalities are marjorly caption-centric. To address this, we leverage the open-source large language model \href{https://huggingface.co/google/flan-t5-xxl}{\texttt{\small{google/flan-t5-xxl}}}~\cite{weifinetuned} 
to automatically generate question-answer pairs for the 3D and audio modalities from their corresponding captions. The process begins by prompting the model with captions to generate potential answers. These answers are then used to prompt the model to generate candidate questions. If the model's response to a question, using the caption as context aligns closely with the initial answer, the example is added to our dataset, yielding $\sim$250k 3D examples from Cap3D~\cite{luo2023scalable}~\footnote{~A subset of 5k point clouds is held-out from Cap3D for the construction of \texttt{DisCRn} (Section \ref{sec:discriminatory_reasoning_dataset}). This exclusion is maintained both in captioning and QA.} and $\sim$24k audio examples from AudioCaps~\cite{kim2019audiocaps}. Details about the data generation and distribution are provided in the supplement. 
\subsection{Discriminative Cross-modal Reasoning}
\label{sec:discriminatory_reasoning_dataset}
\begin{figure}[tb]
\includegraphics[width=\textwidth]{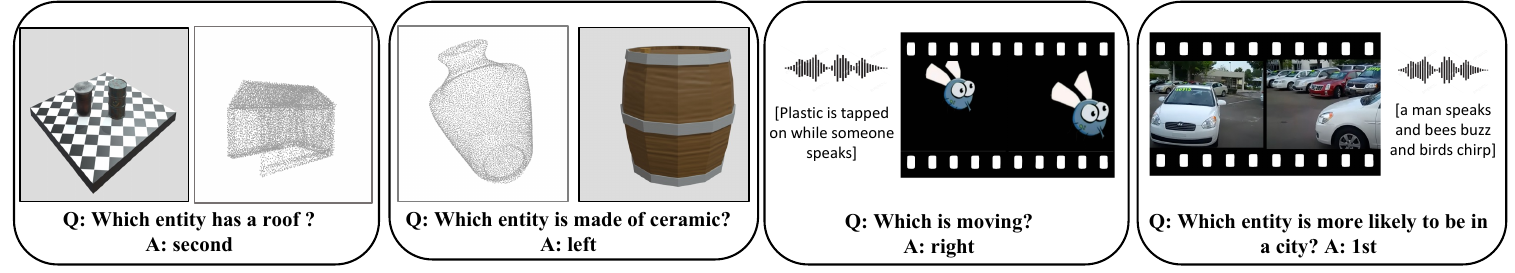}
    \caption{\texttt{DisCRn}. Given two distinct modality inputs, select which one fits the query.}
    \label{fig:discriminatory_reasoning_examples}
\end{figure}

X-InstructBLIP offers a distinct emergent ability: reasoning across different modalities, despite individual modality training. This highlights the model's versatility and potential scalability across numerous modalities. To study this cross-modal reasoning capability, we present a \textbf{Dis}criminatory \textbf{C}ross-modal \textbf{R}easo\textbf{n}ing (\texttt{DisCRn}) challenge dataset. As shown in Figure \ref{fig:discriminatory_reasoning_examples} the task requires the model to discern between the properties of two entities across modalities by selecting which one satisfies a queried property. 
This task mandates the model to not only discriminate the inherent characteristics of the involved modalities but also to consider their relative positioning in the input. This strategic imposition serves to diminish reliance on simplistic text-matching heuristics, order bias, or potential deceptive correlations between modalities.

To generate the dataset, we prompt \texttt{\small{google/flan-t5-xxl}} in a Chain-of-Thought \cite{wei2022chain} manner to generate a set of properties for each dataset instance. Each instance is then paired with a random entity from the dataset to form a (question, answer, explanation) triplet using three examples to leverage in-context-learning~\cite{brown2020language}. A pivotal step in this creation process is a round-trip-consistency check: an example is integrated into the final dataset only when the model's predictions on the generated question, given the captions, exhibits a Levenshtein distance above 0.9 to the example answer. This refined dataset encompasses 8,802 audio-video samples sourced from the AudioCaps validation set, and 29,072 image-point cloud instances from a reserved subset of 5k point clouds from Cap3D~\cite{luo2023scalable}. Each instance in the dataset is coupled with two representations corresponding to the captions: (audio, video) from AudioCaps and (point cloud, images) from Cap3D. Given that the arrangement of the data can be altered, this allows for maintaining a balanced set of answers, not only in terms of the position of the answers, but also the answer modality. Human performance on the task stands at 90\% indicating its high quality. More details found in the supplementary material.  

\section{Experiments}
We study the effectiveness of X-InstructBLIP as a comprehensive solution for incorporating cross-modality into pre-trained frozen LLMs. Following a debrief on the implementation details in Section \ref{sec:implementation}, Section \ref{sec:single_modality} verifies the framework's competitiveness in individual modality-to-text tasks, and explores its emergent cross-modal reasoning ability  even in the absence of joint optimization.

\subsection{Implementation Details}
\label{sec:implementation}
X-InstructBLIP is built on the LAVIS library's framework~\cite{li-etal-2023-lavis} atop of the Vicuna v1.1  7b and 13b models~\cite{chiang2023vicuna}. We adopt {\footnotesize \texttt{EVA-CLIP-ViT-G/14}}~\cite{fang2023eva} as the encoder for image and video, for audio {\footnotesize\texttt{BEATs\textsubscript{iter3+}}}~\cite{pmlr-v202-chen23ag} and for 3D {\footnotesize \texttt{ULIP-2} [\texttt{PointBERT} backbone]}~\cite{salesforceULIP}. 
In the X-Instruct setup, each Q-Former optimizes 188M trainable parameters and learns {\small $K=32$} query tokens with a 
a hidden dimension of size 768 to select a  \textit{single best model} per modality. Raw inputs undergo standardized pre-processing prior to encoding. All Q-Formers are pre-initialized with BLIP-2 stage-1 weights~\cite{dai2023instructblip} except for the video Q-Former which is initialized from the last iteration of the corresponding image Q-Former. 
Details on preprocessing and training hyperparameters for each modality are included in the supplement. The X-LLaVA-style setup linear projection is uniformly initialized and tuned to match the number of trainable parameters in X-Instruct.

All models are optimized on 8 A100 40GB GPUs using AdamW~\cite{loshchilov2018decoupled} with $\beta_1$= 0.9, $\beta_2$= 0.999, and weight decay of 0.05. The learning rate warms up linearly over the initial 1,000 steps from  \(10^{-8}\) to \(10^{-5}\), followed by a cosine decay to a minimum of 0. Evaluation hyper-parameters and templates are consistent across tasks, minimally adapted to each modality as detailed in the supplement.

\subsection{Results}
\label{sec:results}

Our primary aim is to demonstrate the adaptability of our framework across various modalities without relying on large-scale pre-training stages or joint modality data. Nevertheless, to ensure our approach's effectiveness and comparability, we juxtapose its performance to other \emph{methods that employ projections to pre-trained frozen or partially frozen LLMs}, wherever possible. This serves as a mere \textit{sanity check}, verifying that our method is both effective and competitive. 
\subsubsection{Individual Modality Understanding}
\label{sec:single_modality} We evaluate the framework's performance across a range of single modality to text tasks, illustrating its versatility and efficacy across all four explored modalities.
Tables \ref{tab:3d_quantitative_results}, \ref{tab:audio_quantitative_results}, \ref{tab:image_quantitative_results}, and \ref{tab:video_quantitative_results} summarize X-InstructBLIP's out-domain performance across 3D, audio, image, and video. ~\\

\begin{figure}[ht]
     \centering
     \begin{subfigure}[b]{\textwidth}
         \centering
         \includegraphics[width=\textwidth]{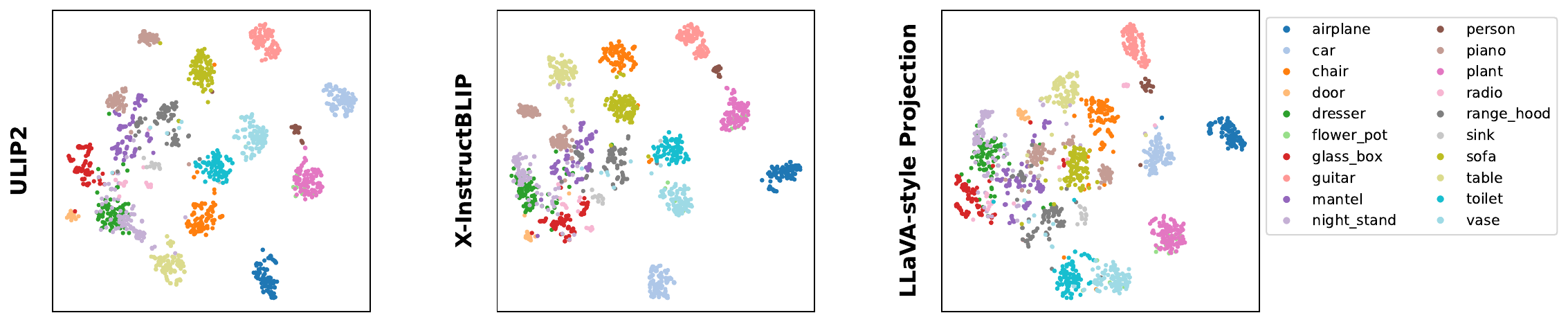}
         \caption{TSne plots of \textit{LLM projections} for 20 randomly sampled Modelnet40 classes.}
         \label{fig:modelnet_a}
     \end{subfigure}
     \hfill
     \begin{subfigure}[b]{0.7\textwidth}
         \centering
         \includegraphics[width=\textwidth]{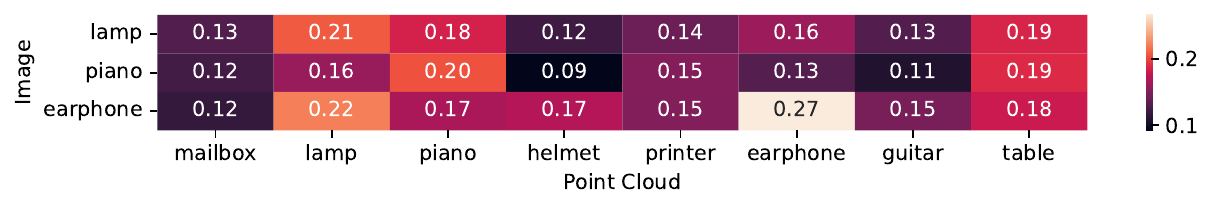}
         \caption{Heatmap of the relative cosine similarity between image and point cloud query outputs from samples in Shapenet.}
         \label{fig:modelnet_b}
     \end{subfigure}
        \caption{Analysis on the alignment of X-InstructBLIP representations.}
\label{fig:modelnet_clusters}
\end{figure}

\noindent\textbf{3D:} Table \ref{tab:3d_quantitative_results} shows the results on zero-shot classification on ModelNet40~\cite{wu20153d} under two setups: classification in closed vocabulary using loss ranking~\cite{li2021align} and open generation where the model is prompted to \texttt{\small describe the 3d model} and correctness is validated if a \textit{single} class from the 40 candidates is present in the generation. In both projection setups, X-InstructBLIP significantly outperforms the InstructBLIP baseline, which processes a single view rendering of the point cloud. Interestingly, the X-Instruct projection setup outperforms not only the X-LLaVA-style projection but also PointLLM~\cite{guo2023point} that learns a similar projection but- unlike this set-up - employs RGB features. It also outperforms PointBindLLM\cite{guo2023point} which trains an adapter on ImageBind\cite{Girdhar_2023_CVPR} image encodings, and relies to the common embedding space to generalize to point clouds, showing the importance of individual modality encoders in our framework. This is further bolstered by the TSNE~\cite{van2008visualizing} visualization of the ULIP-2, X-Instruct and LLaVA-style Projection representations in Figure \ref{fig:modelnet_a} showing that the LLaVA-style Projection breaks class separation leading to lower performance of 16.4 and 19.2 points in classification and open generation accuracy compared to X-Instruct Projection. We further observe, in Figure \ref{fig:modelnet_b},  a mild effect of relative alignment between similar classes across modalities since the cosine similarity of the image and point cloud query outputs of similar classes in Shapenet are higher compared to dissimilar ones. ~\\

\noindent\textbf{Audio:} Table \ref{tab:audio_quantitative_results} shows X-InstructBLIP's performance in audio classification, question answering, and captioning tasks on ESC50 \cite{piczak2015esc}, ClothoAQA \cite{lipping2022clotho}, and Clotho \cite{drossos2020clotho}, respectively. Classification is evaluated both in close (cls) and open generation settings. Both X-InstructBLIP variants outperform ImageBindLLM in all tasks, potentially suggesting that separate encoders and audio specific training data are beneficial for audio-to-text alignment. Notably, X-LLaVA-style Projection outperforms the X-Instruct Projection on Audio QA, while underperforming in all other tasks. This is likely due to the low amount of Audio QA data priming the instruction aware projections to produce a small set of responses.~\\

\noindent\textbf{Image:} While no large variations in performance are expected in comparison to InstructBLIP~\cite{dai2023instructblip}, Table \ref{tab:image_quantitative_results} presents results on image captioning, visual question answering, MME\cite{fu2023mme}, and MMVET\cite{yu2023mm} as a sanity check. While X-InstructBLIP outperforms InstructBLIP on VizWiz~\cite{bigham2010vizwiz} there is a mild drop in performance overall, likely due to the lack of BLIP2 Stage-2 finetuning, and the expanded template space which introduces a trade-off of generalization and performance as shown by the increased prompt robustness of X-InstructBLIP in the supplement.

\noindent\textbf{Silent Video:}  Table \ref{tab:video_quantitative_results} evaluates X-InstructBLIP on out-of-domain video tasks. We compare performance with prominent baselines that rely on \textit{frozen or partially frozen} LLMs and show comparable or improved performance on Video Question Answering (VQA).  However, due to the nature of the instruction aware
\begin{figure}[H] 
\begin{minipage}[H]{0.4\linewidth}
    \fontsize{5pt}{5pt}\selectfont
            \centering
    \begin{tabular}{lll}
        \toprule
            &\multicolumn{1}{c}{Close} &  \multicolumn{1}{c}{Open} \\\midrule
            InstructBLIP~(7b)\cite{dai2023instructblip}  & \textcolor{purple}{31.4} & \textcolor{purple}{23.7}  \\ 
            InstructBLIP~(13b)\cite{dai2023instructblip} & \textcolor{purple}{31.5}& \textcolor{purple}{25.5} \\

            Point-LLMv2+(RGB)~(7b)\cite{xu2023pointllm} & - & \textcolor{purple}{32.3} \\
            Point-LLMv2+(RGB)(13b)\cite{xu2023pointllm} &- & \textcolor{purple}{31.8}\\
                            PointBind-LLM~(7b)\cite{guo2023point} &\textcolor{purple}{47.3} & \textcolor{purple}{36.3} \\ 
  \rowcolor{yellow!10}     X-LLaVA-style Proj.~(7b) &46.4 &30.2 \\  
        \rowcolor{gray!10}        X-Instruct Proj.~(7b) & \textcolor{blue}{62.8}
        & \textcolor{blue}{49.4}
        
        \\ 
      \rowcolor{gray!10}       X-Instruct Proj. (13b) & \textbf{65.1} & \textbf{50.0}\\ \bottomrule 
        \end{tabular}
        \captionof{table}{Zero-shot 3D classification on Modelnet40\cite{uy2019revisiting} test set.}
    \label{tab:3d_quantitative_results}
\end{minipage}\hfill
\begin{minipage}[H]{0.55\linewidth}
    \fontsize{5pt}{5pt}\selectfont
    \centering
   \begin{tabular}{lllll}
\toprule
        &\multicolumn{1}{c}{PT} & MSVD  & VATEX  & MSVDQA \\
        
         & & {\textit{test}}  & {\textit{val} } & {\textit{test}} \\ \midrule
    
        FrozenBiLM\cite{yang2022zero} & $\checkmark$ & - &- &  33.8 \\
        VideoLLaMA\cite{zhang2023video}& $\checkmark$  &-  &- &  \textcolor{blue}{51.6} \\

        InstructBLIP\cite{dai2023instructblip}  &$\times$ & 87.2& 57.6 & 41.2 \\ 
        
  \rowcolor{yellow!10}X-LLaVA-style Proj.~(7b)&$\times$ & 105.3 &46.2 &49.8\\ 

  \rowcolor{gray!10} X-Instruct Proj.~(7b)&$\times$ & \textcolor{blue}{116.1}&\textbf{59.2} & \textbf{51.7}\\ 

 \rowcolor{gray!10} X-Instruct Proj.~(13b) &$\times$ & \textbf{124.3} & \textcolor{blue}{52.0}& 49.2 \\
  \bottomrule \\
    \end{tabular}
    \captionof{table}{Out-Domain Silent Video Results.\\ PT denotes video pretraining stage.}
\label{tab:video_quantitative_results}
\end{minipage}
\vfill
\begin{minipage}[H]{0.95\linewidth}\centering 
\fontsize{5pt}{5pt}\selectfont
            \centering
   \begin{tabular}{llllllll}
\toprule
         &  ESC50\textsubscript{close} & ESC50\textsubscript{open} & ClothoAQA & \multicolumn{2}{c}{Clotho~v1} & \multicolumn{2}{c}{Clotho~v2}\\
       &\verytiny{Acc.} & \verytiny{Acc.} & \verytiny{EM} & \verytiny{CIDEr} &\verytiny{SPIDEr} &\verytiny{CIDEr} &\verytiny{SPIDEr}\\ \midrule
        ImageBind\cite{Girdhar_2023_CVPR} & 66.9& $\times$ & $\times$ &$\times$ & $\times$& $\times$ & $\times$\\\midrule
        MWAFM\cite{li2023multi} & - & - & \underline{22.2} & - & - & - & -\\
        Pengi\cite{deshmukh2023pengi} & - & \textcolor{purple}{53.9} & \underline{64.5} & \textcolor{purple}{\underline{39.6}} & \textcolor{purple}{\underline{30.0}} & \textcolor{purple}{\underline{32.9}} & \underline{27.1}\\
        Kim et. al., 2023\cite{kim2023prefix} &- & - &- &- & - & 19.2& 13.3 \\\midrule
        ImageBind-LLM (7B)\cite{han2023imagebind} & \textcolor{purple}{40.1}& \textcolor{purple}{27.4} & \textcolor{purple}{10.3}\textdagger &\textcolor{purple}{3.7}& \textcolor{purple}{5.5}& \textcolor{purple}{3.2} & \textcolor{purple}{5.5}\\
  \rowcolor{yellow!10}     X-LLaVA-style Proj.~(7b)  &67.4 & 20.3 &  \textbf{26.9} & 25.3& 16.6 &22.0 & 14.8 \\  
\rowcolor{gray!10} X-Instruct Proj.~(7b)  &\textcolor{blue}{75.9} 
&   \textbf{38.2} & 21.4 
& \textbf{29.4} &\textbf{19.5}
& \textcolor{blue}{26.7}  & \textcolor{blue}{17.8}
\\  
 \rowcolor{gray!10} X-Instruct Proj.~(13b) &\textbf{77.1} & \textcolor{blue}{34.6} & \textcolor{blue}{21.7} &\textcolor{blue}{28.7}& \textcolor{blue}{18.8} & \textbf{27.5} & \textbf{18.0} \\
  \bottomrule
    \end{tabular}
    \captionof{table}{Out-Domain Audio Quantitative Results. }
    \label{tab:audio_quantitative_results}
    \end{minipage}\vfill
\begin{minipage}[H]{0.95\linewidth}\centering 
    \fontsize{5pt}{5pt}\selectfont
\centering
\begin{tabular}[width=\textwidth]{llllllll}
\toprule
& &  Flickr30k & \multicolumn{1}{c}{NoCaps}& VizWiz & GQA & MME  & MMVet  \\ \midrule

\multirow{6}{*}{\rotatebox[]{90}{Vision}} &Flamingo~9B\cite{alayrac2022flamingo}&  61.5 & - & 28.8  & - & - &-   \\ 

&BLIP-2\cite{li2023blip}& 76.1  & 107.5 & 29.8& 44.7& \textbf{1293.8}  &22.4 \\ 

 &InstructBLIP \cite{dai2023instructblip}& \textbf{84.5}& \textbf{123.1} & 34.5 &  \textbf{49.5}  &1212.8 &26.2  \\ 
 &MiniGPT4 (7b)\cite{zhu2024minigpt} &-&  -&-& 32.2  &1158.6  &22.1 	\\ 
 
 &LLaVA~(7b)\cite{liu2023visual}&  27.7 &33.1 & -&  -& 717.5&27.4 \\ 

 & LLaMA-adapter~(13b)\cite{zhang2024llamaadapter}&  30.5 &41.7 & -&  -& \textcolor{blue}{1222.0} & -  \\ 
 \midrule
\multirow{5}{*}{\rotatebox[]{90}{X-Modal}}& PandaGPT (13b)\cite{su-etal-2023-pandagpt} &23.0 & 29.7 &- &- & 871.2 &19.6\\ 
& ImagebindLLM (7b)\cite{han2023imagebind}  &23.5&30.4  &-&  - & 989.3 & - \\ 
& \cellcolor{yellow!10} X-LLaVA-style Proj.~(7b) & \cellcolor{yellow!10}6.2 & \cellcolor{yellow!10}22.3 & \cellcolor{yellow!10}27.7&  \cellcolor{yellow!10}41.5 &\cellcolor{yellow!10}866.7 & \cellcolor{yellow!10}17.0  \\
 
&\cellcolor{gray!10} X-Instruct Proj.~(7b) & \cellcolor{gray!10}\textcolor{blue}{82.1} & \cellcolor{gray!10}\textcolor{blue}{117.7} &\cellcolor{gray!10}\textcolor{blue}{34.9} &\cellcolor{gray!10}48.1 &\cellcolor{gray!10}891.8  &\cellcolor{gray!10}\textcolor{blue}{29.0}   \\ 
&\cellcolor{gray!10} X-Instruct Proj.~(13b) &\cellcolor{gray!10}74.7& \cellcolor{gray!10}114.5 &\cellcolor{gray!10}\textbf{36.0}& \cellcolor{gray!10}\textcolor{blue}{49.2} & \cellcolor{gray!10}1174.0 & \cellcolor{gray!10}\textbf{35.1} \\ \bottomrule
    \end{tabular}
\captionof{table}{Out-Domain Image Quantitative Results.}
\label{tab:image_quantitative_results}
\end{minipage}
\caption*{Single Modality Quantitative Results. \underline{Underlined} numbers indicate in-domain evaluations. \textbf{Bold} indicates the top zero-shot performance. \textcolor{blue}{Blue} indicates  second best zero-shot performance. \textcolor{purple}{Purple} denotes evaluations conducted independently. Models denoted with 7b and 13b indicate the underlying LLM size. {\sethlcolor{gray!10}\hl{Gray}} shaded rows correspond to X-InstructBLIP variants, and  {\sethlcolor{yellow!10}\hl{Yellow}} to the LLaVA-style\cite{liu2023visual} model equivalent. CIDEr score ~\cite{vedantam2015cider} is reported for captioning, accompanied by SPIDEr\cite{liu2017improved} score for audio captioning and Top-1 accuracy for QA and classification tasks. \textdagger~ signifies a relaxed exact match metric where the ground truth is a substring of the prediction. }
\end{figure}

\noindent setup, X-InstructBLIP is tuned on other QA tasks, thus having an advantage over VideoLLaMA~\cite{zhang2023video} and FrozenBiLM~\cite{yang2022zero} even though it lacks video pretraining (PT). As we show  in the supplement
, the Video Q-Former component of X-InstructBLIP, initialized with the Image Q-Former's weights, reaches convergence in performance remarkably fast, within about 1,000 iterations. 
For context, VideoLLaMA is pre-trained on the entire WebVideo dataset of 2 million videos, in addition to  BLIP-2 image pre-training. FrozenBiLM, on the other hand, undergoes two epochs of training on the larger WebVideo10M.

\subsubsection{Cross-Modal Joint Reasoning:}
\label{sec:joint}
Despite each modality projection being trained individually, X-InstructBLIP shows strong joint modality reasoning, particularly under the X-Instruct Projection setting. 
Table \ref{tab:joint_audio_video_results} demonstrates X-Instruct's capability to reason jointly over video (V) and audio (A). Notably, X-Instruct Proj.~(7b) is capable of synergizing inputs, displaying an improvement in performance compared to utilizing a single modality when the model is cued with different modalities both in MusicAVQA~\cite{li2022learning} and VATEX~\cite{wang2020vatex}. However, this is not the case for X-LLaVA-style Projection which exhibits the same or lower performance under such a cross-modal setting.

\subsubsection{Cross-Modal Discriminative Reasoning}
\label{sec:discern_eval}
We assess X-InstructBLIP in executing discriminatory reasoning across different modalities using our newly introduced \texttt{DisCRn} benchmark, detailed in Section \ref{sec:discriminatory_reasoning_dataset}. We frame the problem as a realistic open generation problem. The LLM is prefixed with the instruction:
\begin{quote}
{\footnotesize \textcolor{gray}{You are given two inputs. Select exactly one of the two by reference to its relative position (first or second, left or right) that best answers the question.}}
\end{quote}

\begin{figure}[h]
\centering
\begin{minipage}[H]{0.55\linewidth}
\fontsize{6pt}{6pt}\selectfont
    \centering
 \begin{tabular}{lcccccccc}
    \toprule
    &\multicolumn{4}{c}{Music AVQA \textit{test}} &  \multicolumn{4}{c}{VATEX \textit{test}}  \\\midrule
        & A & V & A+V & $\Delta$  & A & V & A+V & $\Delta$ \\
  \rowcolor{yellow!10}  X-LLaVA-style Proj. (7b) &  36.4 & 34.8 & 36.4  & \textcolor{orange}{0.0} & 4.9  &  46.2 &  35.5 & \textcolor{red}{-10.7}\\
\rowcolor{gray!10}   X-Instruct Proj. (7b) &13.4 & 27.2 &  28.1 & \textcolor{ao(english)}{1.3} & 6.7  & 59.2  &\textbf{60.9} &\textcolor{ao(english)}{1.7} \\
 \rowcolor{gray!10}    X-Instruct Proj. (13b) &22.7  & 43.5 & 44.5& \textcolor{ao(english)}{1.0} &  6.1  & 52.0 &58.2  &  \textcolor{ao(english)}{\textbf{6.2}} \\\bottomrule
    \end{tabular}
    \captionof{table}{Emergent Joint (A)udio-(V)ideo Reasoning. $\Delta$ denotes the difference between joint modality and best single modality score.}
\label{tab:joint_audio_video_results}
\end{minipage}\hfill
\begin{minipage}[H]{0.35\linewidth}
    \fontsize{6pt}{6pt}\selectfont
    \begin{tabular}{lllll}
    \toprule
    & A-V & Img-3D \\\midrule
     Caption Baseline (7b) &30.8  & 41.8\\ 
  \rowcolor{yellow!10}    X-LLaVA-style Proj. (7b) & \textbf{47.1}& 41.4 \\
  \rowcolor{gray!10}    X-Instruct Proj.~(7b) & 34.0 & 48.1\\
 \rowcolor{gray!10}     X-Instruct Proj.~(13b) & 45.5 &\textbf{48.8}    \\\bottomrule
    \end{tabular}
    \captionof{table}{\texttt{DisCRN} evaluation.\\\qquad\\\quad }
    \label{tab:disc_dataset_results}
\end{minipage}
\vfill
\end{figure}

\noindent In prompting X-Instruct Proj. (7b) we found that using a Q-Former captioning prompt different from the comparative prompt provided to the LLM model induces a more general representation that was more applicable for the comparative task, as such we employ this approach for the results in Table \ref{tab:disc_dataset_results}. This is likely due to the lack of comparative data in fine-tuning since each modality Q-Former is trained separately. Future work can explore the effect of different prompts conditioned on different parameters in the instruction-aware training setup (e.g. data, templates, joint training, and LLM partial or full optimization). For the video-audio comparison, we select two frames for each modality to allow for a more balanced generation influence. 

To benchmark our model's capabilities, we incorporate a robust captioning baseline by substituting the query outputs with captions corresponding to the modalities using the Vicuna 7b model.
For images, 3D, and video modalities, we elicit captions by prompting InstructBLIP~\cite{dai2023instructblip} to {\texttt{\small Describe the image/video}}. For 3D, a randomly chosen rendering view of the point cloud is provided to InstructBLIP. For video we follow \cite{dai2023instructblip} and sample four frames and concatenate their output representations as input to the model. For audio we use WavCaps~\cite{wavcaps}.

While the X-InstructBLIP framework produces models that outpefrom the strong captioning baseline by a significant margin, there is no conclusive remark on which of the two projection types is more suitable for cross-modal discriminative reasoning\footnote{It is worth noting that using a small sub-sample of the data we observed that the task is prompt sensitive, mainly in the language only setting. We leave it to future work to systematically evaluate the model's ability on the task based on different prompts and in-context examples.}. X-LLaVA Proj. outperforms X-Instruct Proj. on Audio-Video, likely due to its stronger Audio QA performance also reported in Table \ref{tab:audio_quantitative_results}. For image-3D the opposite is true, signifying the intuitive result that the individual modality performance plays a role in cross-modal reasoning abilities. It is worth noting, however, that X-Instruct Proj. exhibits the ability to switch from discriminative to joint reasoning, by either discriminating or combining the inputs to generate a response. As seen in Table \ref{tab:joint_audio_video_results} this is not the case for X-LLaVA style projections, suggesting that the instruction aware representations might prime the LLM to respond more aptly to the task in question. 

\subsection{Ablations}
\noindent\textbf{Prefix Effect:} We explore the effect of prefixing the modality input with a modality specific prefix in Table \ref{tab:prefix_effect}. We compare performance of X-Instruct Proj. (7b) with  X-Instruct Proj.\textsubscript{no-prefix} which is trained similarly to X-Instruct Proj. (7b), with the distinction that the modality type is not prepended to the modality's LLM input tokens before feeding into the LLM for training and inference. In both audio and 3D single modality tasks removing the prefix consistently hurts the performance. This improvement is likely due to the fact that the Q-Former is relieved from the extra burden to encode the type of modality and instead reserves  bandwidth for semantic information. Including the prefix also allows the model to learn to combine modalities better as shown by the improved performance over the single modality for MusicAVQA and VATEX.  Initially, it was theorized that the model's inability to differentiate tokens corresponding to each modality, treating them instead as a continuous stream, might be the cause for this behavior. However, the results from the image-3D cross-modal reasoning task where the prefix-less model outperforms the prefixed one by 10 points challenge this view. It appears that the inclusion of cues may be prompting the model to encode modality-specific information, which is beneficial in joint reasoning scenarios. This specialized encoding does not, however, prime the model to recognize and process characteristics usually associated with other modalities, required for enhanced performance in contrastive tasks. The underlying rationale is that the language model, already tuned to generate modality-relevant outputs, leads the Q-Former to primarily receive feedback on modality-specific generation during training, also accounting for the improvements in single modality.   
\begin{table}[tb]
    \centering
    \fontsize{5pt}{5pt}\selectfont
    \begin{tabular}{cccc}
    \toprule
   Modality & Task  & X-Instruct Proj. & X-Instruct Proj.\textsubscript{no-prefix} \\ 
    \midrule
   \multirow{2}{*}{3D} & Modelnet40 \textit{close}  & \textbf{62.8} & 60.9 \\
                      &  Modelnet40 \textit{open} & \textbf{49.4} & 46.7 \\\midrule
    \multirow{5}{*}{Audio} & ESC50 \textit{close}  & \textbf{75.9} & 67.5 \\
                      &  ESC50 \textit{open} & \textbf{38.2} & 36.0\\
                    &  ClothoAQA & \textbf{15.4} &9.9 \\
                    &  Clotho v1/v2 & \textbf{29.4}/\textbf{26.7} & 26.9/24.5\\\midrule
 \multirow{3}{*}{Audio+Video} & MusicAVQA (A/V/A+V/$\Delta$) & \textbf{13.4}/27.2/\textbf{28.1}/\textcolor{ao(english)}{1.3} & 8.9/\textbf{27.3}/22.3/\textcolor{red}{-5.0}\\
 & VATEX (A/V/A+V/$\Delta$) & 6.7/59.3/\textbf{60.9}/\textcolor{ao(english)}{1.7}& \textbf{6.8}/\textbf{59.5}/58.3/\textcolor{red}{-1.2} \\
 & DisCRn & \textbf{34.0} & 31.4 \\ \midrule
Image+3D & DisCRn &48.1 & \textbf{57.7}   \\            
     \bottomrule
    \end{tabular}
    \caption{Ablation: Prefix Effect}
    \label{tab:prefix_effect}
        \centering
   \begin{tabular}{llllll}
\toprule
      & ESC50\textsubscript{close} & ESC50\textsubscript{open} & ClothoAQA & \multicolumn{1}{c}{Clotho~v1} & \multicolumn{1}{c}{Clotho~v2}\\ \midrule
        \rowcolor{gray!10} X-Instruct Proj.~(7b)  & \textbf{75.9} &   \textbf{38.2} & \textbf{15.4}& \textbf{29.4} & {26.7}\\
     \rowcolor{gray!10} X-Instruct Proj.~(7b)\textsubscript{no-init}  & 70.0  &{37.8} & 11.9 &   {29.3} & \textbf{27.4}\\    \bottomrule
    \end{tabular}
    \captionof{table}{Out-Domain Audio Quantitative Results.}
    \label{tab:blip_init}
\end{table}

\noindent\textbf{BLIP-2 Initialization} We also explore the effectiveness of the BLIP-2 initialization by training the audio Q-Former in X-Instruct Proj. (7b) using a random initialization approach denoted as X-Instruct Proj.~(7b)\textsubscript{no-init}. Table \ref{tab:blip_init} demonstrates the benefits of this prior, indicating that it's possible to integrate new modalities into our framework without extensive pre-training, since from the modalities considered, audio is the least likely to benefit from image-text pre-training. Future research should delve into the effects of modality-specific pre-training, as they are outside of our scope. The most significant improvement is observed in question answering, indicating that BLIP-2 weight initialization appears to enhance instruction awareness more than direct audio-language alignment, corroborated by the gap in closed vocabulary classification performance.

\section{Conclusion}
This study introduces X-InstuctBLIP, a scalable framework for independently aligning the representation of multiple modalities to that of a frozen LLM demonstrating competitive results compared to leading  methods across all addressed modalities. The framework exhibits emergent cross-modal reasoning, despite separate modality training. To test this emergent ability a new cross-modal discriminatory reasoning task \texttt{DisCRn} is introduced to show that the framework yields models that can outperform strong captioning baselines across all four examined modalities. Despite the effectiveness of the method, the task remains an open challenge.
We also find complexities and unanswered questions within each modality, paving the way for future explorations across and within modalities.

\clearpage  

\title{Supplementary Material for \texttt{X-InstructBLIP}: A Framework for Aligning Image, 3D, Audio, Video to LLMs and its Emergent Cross-modal Reasoning}
\author{Artemis Panagopoulou\inst{1}\thanks{Work done while interning at Salesforce Research} \and
Le Xue\inst{2,\ast\ast} \and
Ning Yu\inst{2, \ast\ast} \and Junnan Li\inst{2} \and Dongxu Li\inst{2} \and Shafiq Joty\inst{2} \and Ran Xu\inst{2} \and Silvio Savarese\inst{2} \and Caiming Xiong\inst{2} \and Juan Carlos Niebles\inst{2}}

\authorrunning{A.~Panagopoulou et al.}
\titlerunning{\texttt{X-InstructBLIP}}

\institute{University of Pennsylvania \\
 \email{artemisp@seas.upenn.edu}\and
Salesforce AI Research \\}
\maketitle

\section{Data Generation}

\label{app:data_gen}
\renewcommand{\thefootnote}{\fnsymbol{footnote}}
\footnotetext[2]{Equal mentorship contribution.}
\setcounter{footnote}{0}
\renewcommand{\thefootnote}{\arabic{footnote}}
\subsection{Instruction Tuning Data Augmentation}
\label{app:mm_qa_data_gen}

\begin{figure}[tb]
    \centering
    \includegraphics[width=\textwidth]{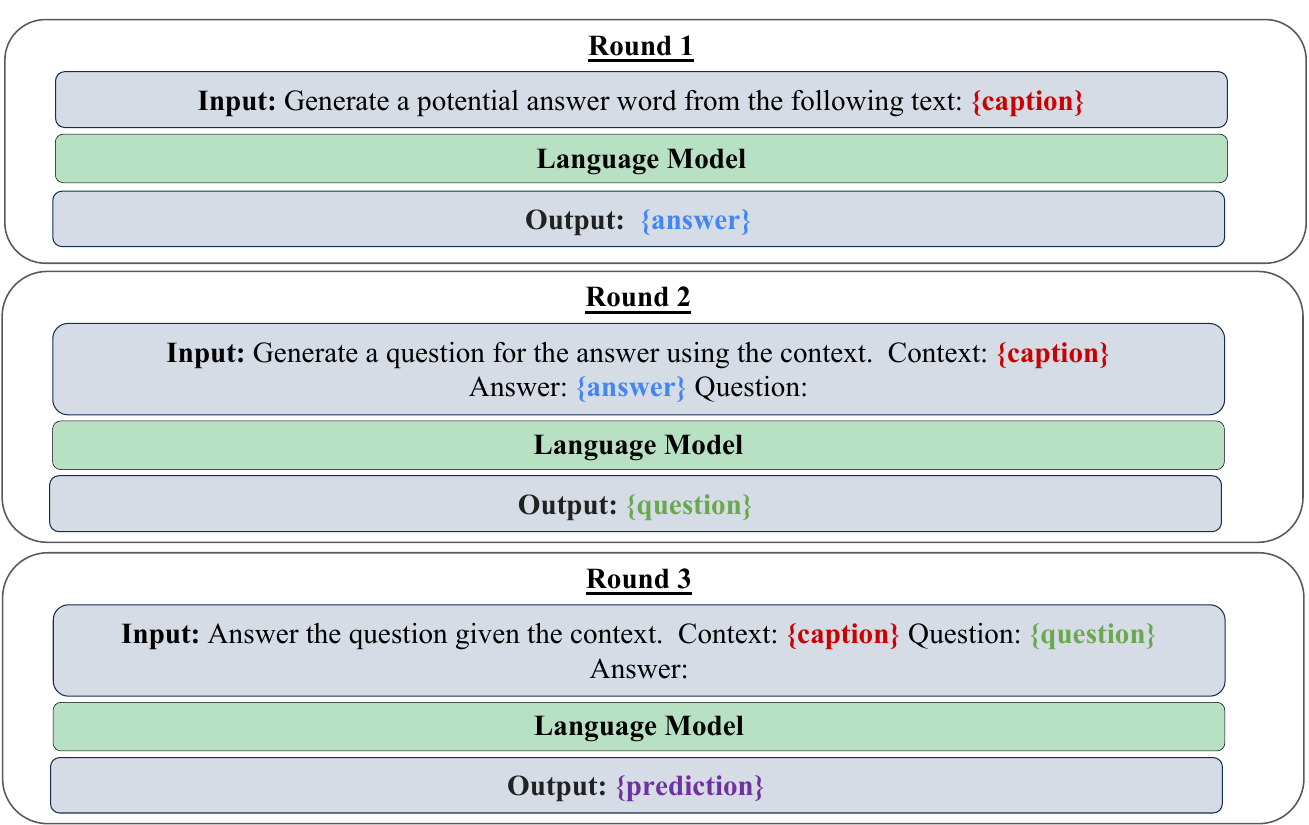}
\caption{Round-Trip-Consistency Prompting for QA Datasets in 3D and Audio.}

    \label{fig:round_trip_consistency}
\end{figure}

For the audio and 3D modalities, the available range of tasks for instruction tuning is relatively limited. To address this challenge we follow a common paradigm in the literature~\cite{xu2017video} and extract question-answer pairs from captioning datasets, specifically from captions consisting of 10 words or more. Figure \ref{fig:round_trip_consistency} delineates the procedure to automatically generate question answering data from captioning datasets. The \href{https://huggingface.co/google/flan-t5-xxl}{\texttt{\small{google/flan-t5-xxl}}} model from \href{https://huggingface.co/docs/transformers/index}{huggingface-transformers} is employed, and is prompted to produce candidate single-word answers based on the caption. 
Subsequently, the model is tasked with generating a relevant question using the answer and context as inputs. The method of round-trip-consistency~\cite{paranjape2021retrieval} is utilized to retain only those question-answer pairs that align with the context. This alignment is verified by ensuring that the Levenshtein partial similarity between the predicted and initial answers is greater than 0.90, calculated using the \href{https://pypi.org/project/fuzzywuzzy/}{Fuzzy Wuzzy} Python package. Subsequently, we apply a string matching post-processing to filter out instances that do not conform to the prescribed  format. As a result, 250,070/1,157 suitable training/validation examples are derived from an initial 661,576/5,000 3D-caption samples from Cap3D~\cite{luo2023scalable}, and 24,156/1,653 training/validation examples are derived from 38,695/1,900 original audio-caption samples from AudioCaps~\cite{kim2019audiocaps}.  Moreover, for 3D data, it is imperative to ensure that the question-answer pairs do not allude to color. This is due to the fact that the 3D encoder does not capture color characteristics. To achieve this, the language model is directed to reformulate the captions by omitting any references to color, prompted as: \texttt{\small Rewrite the sentence \textcolor{red}{\{caption\}} by eliminating any color mentions}, prior to implementing the round-trip-consistency check. A short human evaluation on 50 samples for each modality shows that 90\% of the generated audio and 82\% of the 3D data is correct. Table \ref{tab:qa_examples} presents a random sample of the generated data and table \ref{table:qa_gen_data_statistics} provides an overview of the datasets's distribution statistics. It is worth noting that the error cases are typically due to non-sensical questions, rather than wrong answers. For example the following pairs were marked as non-sensical: \textit{What is the sewing machine running at? speed}, \textit{What does the steam whistle do? hisses}, \textit{What is the 3D model of a brick wall with holes and stacked cubes, resembling? elements}, and \textit{What is the hat with? pattern}.

\begin{table}[tb]
\small
\centering
\fontsize{7pt}{8pt}\selectfont
\begin{tabular}{l|p{7cm}p{3cm}l} 
\toprule
 & \textbf{Caption}                                                & \textbf{Question}                            & \textbf{Answer} \\ \noalign{\hrule height .3pt}
\multirow{4}{*}{\rotatebox[]{90}{Audio}   }& A woman speaks while types a keyboard; &  What is the woman typing on?   & Keyboard        \\\addlinespace[-.4ex]\cmidrule{2-4}\addlinespace[-.4ex]
&
 A man are talking while multiple dogs are barking around them;& What is the dog doing?                       &  Barking         \\\addlinespace[-.4ex]\cmidrule{2-4}\addlinespace[-.4ex]
& A man speaks and a crowd applauds, he continues talking;     &   What does the crowd do after the man speaks? &  Applauds        \\\addlinespace[-.4ex]\cmidrule{2-4}\addlinespace[-.4ex]
& A plane flies in the distance as a man speaks and metal clinks.
&  What does the metal do?                      & Clinks         \\ \noalign{\hrule height .8pt}
\multirow{3}{*}{\rotatebox[]{90}{3D} } &
A 3D model of a wooden chair and stool with a chained bucket on it & What is on the stool? &  Bucket \\\addlinespace[-.4ex]\cmidrule{2-4}\addlinespace[-.4ex]
& A 3D model of a moss-covered stone,  resembling a leaf, paper map, and rock & What is covering the stone? &  Moss \\\addlinespace[-.4ex]\cmidrule{2-4}\addlinespace[-.4ex]
&A balloon with a string attached,  featuring a teddy bear and a cat face on it  &  What is the object with a string attached? &  Balloon \\\addlinespace[-.4ex]\cmidrule{2-4}\addlinespace[-.4ex]
& A 3D model of various food items,including an oyster, a piece of fruit,  and different forms of eggs. &  What is the food item that is a shellfish?   & oyster \\\bottomrule
\end{tabular}
\caption{Automatically Generated QA examples from Captioning Data.}
 
\label{tab:qa_examples}
\end{table}

\begin{table}[tb]
\centering
\fontsize{7pt}{8pt}\selectfont
\begin{tabular}{ccccc}
\toprule
\textbf{Dataset} & \multicolumn{2}{c}{\texttt{AudioCapsQA}} &  \multicolumn{2}{c}{\texttt{Cap3DQA}} \\
& \textit{train}& \textit{val} & \textit{train} &  \textit{val} \\
\midrule
Size  & 24,156 & 1,274 &250,070 & 1,157\\
\midrule
Distinct Questions  & 10,010 & 810& 67,001 & 953 \\
\midrule
Distinct Answers  & 1,636 & 374 & 4,555 & 451 \\
\midrule
Avg. Question Length (words) & 6.0  & 6.1  & 6.8  & 7.0  \\
\midrule
Vocabulary Size & 2,951  & 723  & 12,771  & 1,022\\
\bottomrule
\end{tabular}
\caption{QA Generated Dataset Statistics}
\label{table:qa_gen_data_statistics}

\end{table}

\subsection{Cross-modal Discriminative Reasoning Data Generation}
\label{app:disc_dataset}

To assess the cross-modal reasoning capabilities of X-InstructBLIP, we devised a unique task that repurposes existing captioning datasets, specifically focusing on data representable in multiple modalities. We chose the AudioCaps~\cite{kim2019audiocaps} validation dataset and reserved a subset of 5k examples from Cap3D~\cite{luo2023scalable} as our validation dataset, ensuring that the 3D projection is not exposed to this subset during the training phase in either captioning or 3DQA settings.

The audio data from AudioCaps originates from \href{https://www.youtube.com/}{Youtube} videos, allowing us to download the corresponding video files using their YouTube IDs. For Cap3D, we employed the associated point clouds and randomly selected one rendered image from the available eight view angles.

\begin{figure}[tb]
    \centering
    \includegraphics[width=\textwidth]{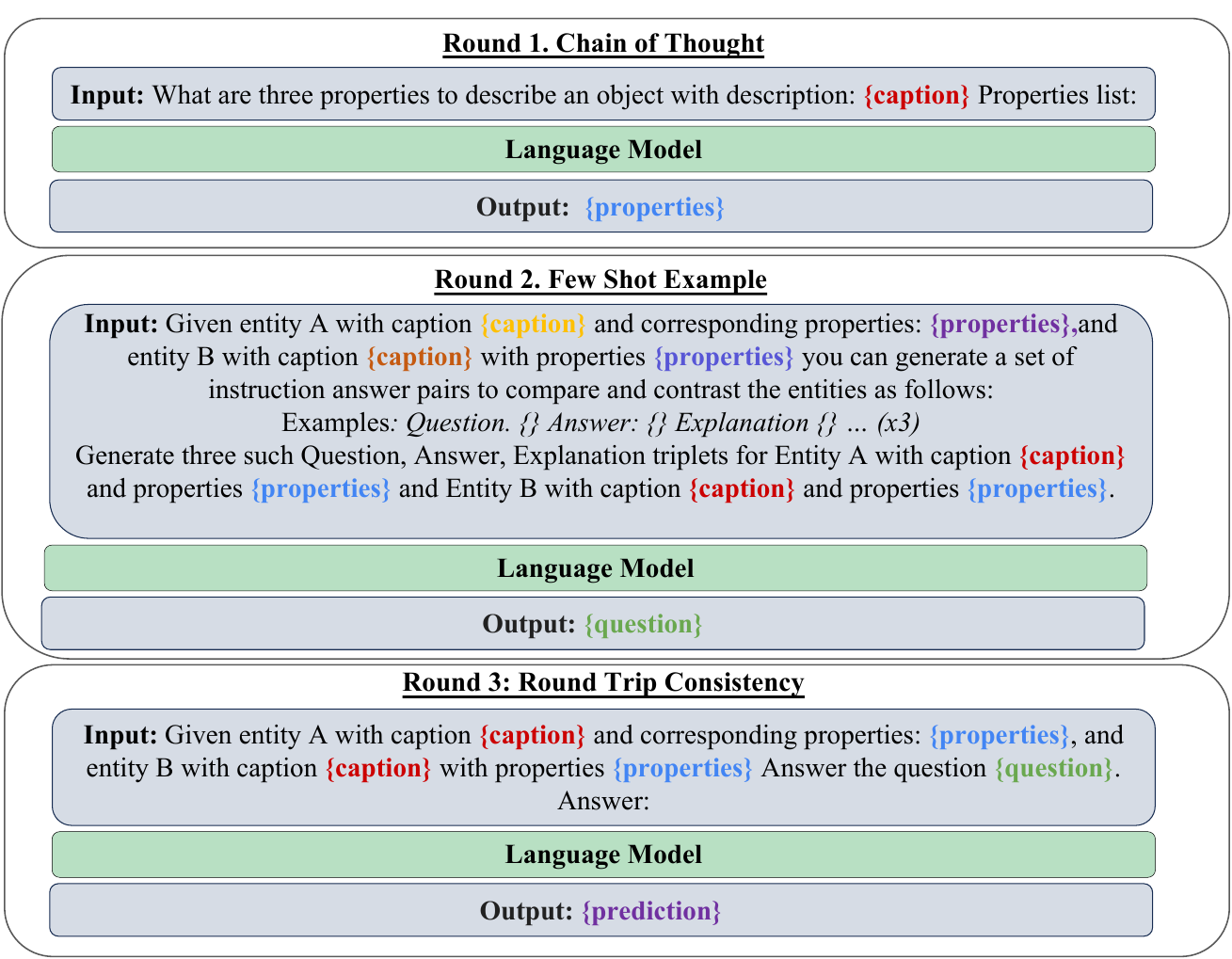}

    \caption{Cross-modal Discriminative Reasoning Dataset Generation Framework}
    \label{fig:disc_dataset_generation}
\end{figure}

A depiction of the data generation procedure, also outlined in the main text, is provided in Figure \ref{fig:disc_dataset_generation}. During the evaluation, we maintain a balance, ensuring each option (A or B) serves as the ground truth 50\% of the time. Given that this problem is structured as an open vocabulary generation task, we expanded the ground truth answer space to include synonyms and equivalent expressions, such as \texttt{\footnotesize[\{answer modality\}, left, 1st, 1, first, input 1, entity 1, object 1, input A, entity A, object A]} and \texttt{\footnotesize[\{answer modality\}, right, 2nd, second, input 2, entity 2, object 2, input B, entity B, object B]}, corresponding to whether the first or the second input is the ground truth.  The human performance on a subsample of 100 examples of the dataset is 90\%.  Table \ref{table:disc_gen_data_statistics} provides an overview of the datasets's distribution statistics.

\begin{table}[tb]
\centering
\small
\fontsize{7pt}{8pt}\selectfont
\begin{tabular}{ccc}
\toprule
\textbf{Dataset} & Audio-Video & Video-3D \\
\midrule
Total Size  & 8,802 & 28,173\\
\midrule
Number of Distinct Questions  & 1,212 & 3,100\\
\midrule
Average Question Length  & 5.8 words & 6.6 words\\
\midrule
Question Vocabulary Size & 701 words & 1,272 words\\
\midrule
\end{tabular}
\caption{\texttt{DisCRn}: Discriminative Cross-modal Reasoning Dataset Statistics}
\label{table:disc_gen_data_statistics}
\end{table}

\section{Video Q-Former Fine-Tuning Versus Image Initialization}
\label{app:video_training_ablation}

To explore the impact of further training the Image Q-Formers on video data, Table \ref{tab:vid_tuning_ablation} presents the results of evaluating video tasks using the weights from the Image Q-Formers. It is evident that training on video data enhances performance. However, it's worth noting that the Video Q-Formers reach convergence at an earlier stage (15k and 5k iterations for Vicuna7b and Vicuna13b, respectively). This is likely because the Q-Formers have already achieved semantic understanding during the image alignment phase, requiring minimal additional training to capture the nuances of sequential video projections. The higher drop in performance in MSVD~\cite{chen2011collecting} captioning compared to VATEX~\cite{wang2020vatex} is likely due to the closer similarity between MSVD and the held-in MSRVTT~\cite{xu2016msr} dataset distribution. There is a notably lower drop in performance for Video QA tasks, owing to the more constraint nature of the task - training on videos would not significantly increase the performance since the answer is typically constrained in one frame~\cite{buch2022revisiting}, and as such processing that frame would be almost equivalent to processing it in the image. The improvement probably stems from identifying the answer across a longer sequence of query tokens.
\begin{table}[hb]
\fontsize{7pt}{8pt}\selectfont
\centering

   \begin{tabular}{llll}
\toprule
        & MSVD  & VATEX  & MSVD QA  \\
        
        & {\tiny\textit{test}}  & {\tiny\textit{val} } & \textit{test} \\\noalign{\hrule height .3pt}
         \rowcolor{yellow!10} X-LLaVA Style Proj. (7b) &105.3 &46.2 &49.8 \\\noalign{\hrule height .3pt}
          \rowcolor{yellow!10} X-LLaVA Style Proj. (7b){\tiny[\textit{image}]} &16.4
 & 10.7&  23.2\\\noalign{\hrule height .3pt}
  \rowcolor{gray!10} X-Instruct Proj. (7b) & 116.1&\textbf{59.2} & \textbf{51.7} \\\noalign{\hrule height .3pt}
  
  \rowcolor{gray!10} X-Instruct Proj. (7b)~{\tiny[\textit{image}]} & 42.4\verytiny{($\downarrow$73.7)} & 28.1\verytiny{($\downarrow$30.1)}& 39.7\verytiny{($\downarrow$12.0)} \\\noalign{\hrule height .3pt}

   \rowcolor{gray!10} X-Instruct Proj.$_\text{no-prefix}$ (7b){\tiny[\textit{image}]}&
  62.0\verytiny{($\downarrow$56.7)}&52.6\verytiny{($\downarrow$6.9)}& 38.8\verytiny{($\downarrow$11.7)}\\ \noalign{\hrule height .3pt}
 
 \rowcolor{gray!10} X-Instruct Proj. (13b) & \textbf{124.3} & 52.0& 49.2 \\\noalign{\hrule height .3pt}
   \rowcolor{gray!10} X-Instruct Proj. (13b)~{\tiny[\textit{image}]} & 78.7\verytiny{($\downarrow$45.6)} &53.5\verytiny{($\uparrow$1.5)} & 36.0\verytiny{($\downarrow$13.2)} 
  \\ \bottomrule
    \end{tabular}
    \caption{Impact of Training Image Q-Formers on Video. Models labeled [\textit{image}] utilize the Image Q-Former for video alignment.}
 \label{tab:vid_tuning_ablation}
\end{table}

\section{In-Domain Evaluations}

Table \ref{tab:in_domain_eval} presents in-domain performance for a sample of datasets seen in training across all four modalities. It's important to clarify that when we refer to `in-domain,' we are specifically referring to datasets that were sampled during the training process. However, it's crucial to note that this does not constitute explicit fine-tuning, as there is no guarantee that the Q-Former has encountered the entirety of the dataset during its training.

\section{Prompt Robustness}
\label{app:prompt_robustness}

Table \ref{tab:prompt_robustness} compares performance between InstructBLIP (7b) and X-Instruct Proj. (7b) on NoCaps~\cite{agrawal2019nocaps}, using prompts not encountered in the optimization of either model. While X-InstructBLIP exhibits some performance variability, it maintains more than half the standard deviation of InstructBLIP. This variance can be attributed to the expanded vocabulary in our templates, allowing the Q-Former to better associate an instruction with a specific task. For example, in the case of prompt P2: \textit{``Provide a recap of what is happening in the picture"}, InstructBLIP maintains high performance as it closely resembles an in-domain prompt \textit{``Use a few words to illustrate what is happening in the picture"}. Note that the performance drop in InstructBLIP is mostly attributed to the language model resorting to generating longer descriptions when the Q-Former outputs have not captured the task, resulting in hallucinations in later stages of generation.
\begin{table}[h]
\fontsize{7pt}{7pt}\selectfont
\centering
\begin{tabular}{lll}
\toprule
 & InstructBLIP (7b) &\cellcolor{gray!10}  X-Instruct Proj. (7b) \\
\midrule
P1 & 1.0 & \cellcolor{gray!10} \textbf{88.0} \\
P2 & \textbf{121.9} & \cellcolor{gray!10} 109.7 \\
P3 & 0.9 &\cellcolor{gray!10}  \textbf{54.9} \\
P4 & 5.4 &\cellcolor{gray!10}  \textbf{112.7} \\
P5 & 0.8 & \cellcolor{gray!10} \textbf{111.5} \\
Avg & 26.3 &\cellcolor{gray!10}  \textbf{83.0} \\
Std. & 43.8 & \cellcolor{gray!10} \textbf{20.8} \\
\bottomrule
\end{tabular}
\begin{tabular}{ll}
&\\
P1 & In a few words describe the basic features of this image.\\
P2 & Provide a recap of what is happening in the picture. \\
P3 & I'd like to hear your interpretation of this image. What do you see? \\
P4 & Provide a verbal snapshot of what's happening in this image. \\
P5 & Please articulate the elements and context of this image \\
\end{tabular}
\caption{Robustness to unseen prompts on NoCaps {\small (\textit{val})}~\cite{agrawal2019nocaps}.}
\label{tab:prompt_robustness}
\end{table}

\begin{table}[hb]
\fontsize{7pt}{7pt}\selectfont
\centering
   \begin{tabular}{lllllllllllll}
\toprule

 &\multicolumn{3}{c}{Image}  &\multicolumn{2}{c}{3D}  &\multicolumn{4}{c}{Video }& \multicolumn{3}{c}{Audio} \\
&OKVQA & \multicolumn{2}{c}{COCO}  &  \multicolumn{2}{c}{Cap3D} & \multicolumn{2}{c}{MSRVTT}    & \multicolumn{2}{c}{MSRVTT QA} & \multicolumn{3}{c}{AudioCaps} \\
& {\tiny\textit{test}} &  {\tiny\textit{val} }& {\tiny\textit{test} }& {\tiny\textit{val}} & {\tiny\textit{qa-val} } & {\tiny\textit{val}} &  {\tiny\textit{test} }&  {\tiny\textit{val}} &  {\tiny\textit{test} }&  {\tiny\textit{val}} &  {\tiny\textit{test}} & {\tiny\textit{qa-val}} \\\noalign{\hrule height .3pt}
 \multirow{2}{*}{Finetuned SOTA} &66.1&-&155.1&-&-&-&80.3&-&48.0~&-&78.1
   &-\\  &\multicolumn{2}{c}{\tiny\cite{pmlr-v202-driess23a}} &\multicolumn{2}{c}{\tiny\cite{li-etal-2022-mplug}}&&&\multicolumn{2}{c}{\tiny\cite{xu2023mplug2}}&{\tiny\cite{xu2023mplug2}}&\multicolumn{2}{c}{\tiny\cite{chen2023vast}}
   &\\\noalign{\hrule height .3pt}
       InstructBLIP~(T5xl) &{48.6}&{137.7} &{140.2}&-&-&{44.1}& {44.0}& 25.0&{22.3}& -&-&-\\\noalign{\hrule height .3pt}
        InstructBLIP~(T5xxl) &{47.8}&{139.1}&{140.8}&-&-&{41.5}&{47.8}&25.6&{21.4}&-&-&- \\\noalign{\hrule height .3pt}
        InstructBLIP~(7b)&{57.3}& {141.0}&{142.3}&-&-&{28.1}&{31.1}&22.1&{18.7}&-& -&-  \\\noalign{\hrule height .3pt}
        InstructBLIP~(13b) &{56.3}&{139.1} &{141.0}&-&-&{36.7}&{37.1}&24.8&{20.2}&-& -&- \\\noalign{\hrule height .3pt}
  \rowcolor{yellow!10} X-LLaVA Style Proj. (7b) & 28.5& 126.0 &  118.1&126.7&39.9&55.5&53.1&41.0&41.4&44.3&46.1&53.2\\\noalign{\hrule height .3pt}
  \rowcolor{gray!10} X-Instruct Proj. (7b) & 52.5&137.7& 138.2&142.1&53.6&61.0&57.6&44.6&42.1&44.6& 67.9&41.2\\\noalign{\hrule height .3pt}

 \rowcolor{gray!10} X-Instruct Proj. (13b) &51.9&128.2& 128.7&148.8&54.9&57.7&52.2&36.4&36.1&54.2 & 53.7&37.4\\ \bottomrule
    \end{tabular}
    \caption{In-Domain performance across modalities.} 
 \label{tab:in_domain_eval}
\end{table}

\section{Training Details}
\label{app:implementation}

Prior to encoding, raw inputs undergo standardized pre-processing: images are resized to \(224 \times 224\) resolution with random cropping and normalization; audio files undergo mono conversion and filter bank pre-processing followed by normalization as in \cite{pmlr-v202-chen23ag} over two 5-second frames; videos are uniformly sampled to 5 frames subject to the same pre-processing as images, and 3D point clouds are uniformly  sampled and normalized to 8k points as in \cite{salesforceULIP, xue2023ulip}. All modality Q-Formers are pre-initialized with BLIP-2~\cite{dai2023instructblip} stage-1 weights except for the video Q-Former which is initialized from the last iteration of the corresponding image Q-Former and optimized for 15k/5k steps for the Vicuna 7b and 13b models respectively. 

Table \ref{tab:implementation_param} compiles the training hyperparameters employed for each modality and model. The X-Instruct Proj. $_\text{no-prefix}$ variant is trained similarly to X-Instruct Proj., with the notable distinction that the modality type is not prepended to the modality's query outputs, both during training and inference. Following \cite{dai2023instructblip} that noted that sampling ratios play an important role in training we perform some minor modifications in the sampling ratios that we show in tables \ref{tab:coco_uspample} and \ref{tab:upsample_msrvtt} are effective in improving performance. The decisions are discussed further below. It is worth noting that due to the large amount of experiments consisting of all modalities, we did not exhaust all possibilities, and there may be better training configurations. We leave this to future work to be explored.  

As each modality exhibits unique characteristics, we have customized the training approach for each one. For instance, the 3D and Audio projections are trained for the maximum number of iterations specified in Table \ref{tab:implementation_param}.

The Vicuna7b Image projection undergoes training for 735k iterations, utilizing normalized data sampling. Additionally, an extra 40k iterations are performed with the sampling ratio of COCO Captions~\cite{changpinyo2021conceptual} set to 3.0 while keeping the other ratios consistent with the original sampling. This adjustment leverages the clean annotations of COCO Captions, mitigating noise introduced by larger image datasets. However, this upsampling technique is not applied to the Vicuna13b Image Q-Former, since it appears to lower out of distribution performance in non-captioning tasks as shown in table \ref{tab:coco_uspample}. It could be that due to the smaller batch size, Vicuna13b is less sensitive to noisy data, since it effectively sees less of them. In both cases, the last checkpoint from the iterations specified in Table \ref{tab:implementation_param} is chosen, with guidance from the COCO Captions validation dataset. Note that we optimize the Image Q-Former for 10 times more iterations that InstructBLIP. The reason is that we maintain conformity with the other Q-Formers and do not intitialize the cross-attention layers from BLIP-2 pretraining nor we allow for stage-2 training. Nevertheless, we show that with enough iterations, the cross attention layers can be learned equivalently without the need of the contrastive auxiliary losses of BLIP-2 nor stage-2 training. 

The Vicuna7b video projections are initialized from the best Vicuna7b image projection and undergoes validation every 5k iterations on the MSRVTT captioning~\cite{xu2016msr} dataset. The selection process involves choosing the checkpoint that precedes any drop in performance during the subsequent validation rounds even if there is a better performing checkpoint later on in training, to avoid overfitting to the MSRVTT skeletal captions. Table \ref{tab:upsample_msrvtt} quantitatively shows our observations. Due to the initialization of the video Q-Former with the well trained image Q-Former, the noisy captions of WebVid2M reduce the performance instead of improving it. However, this is corrected with cleaner data.

\begin{figure}[h]
    \centering
    \includegraphics[width=.5\textwidth]{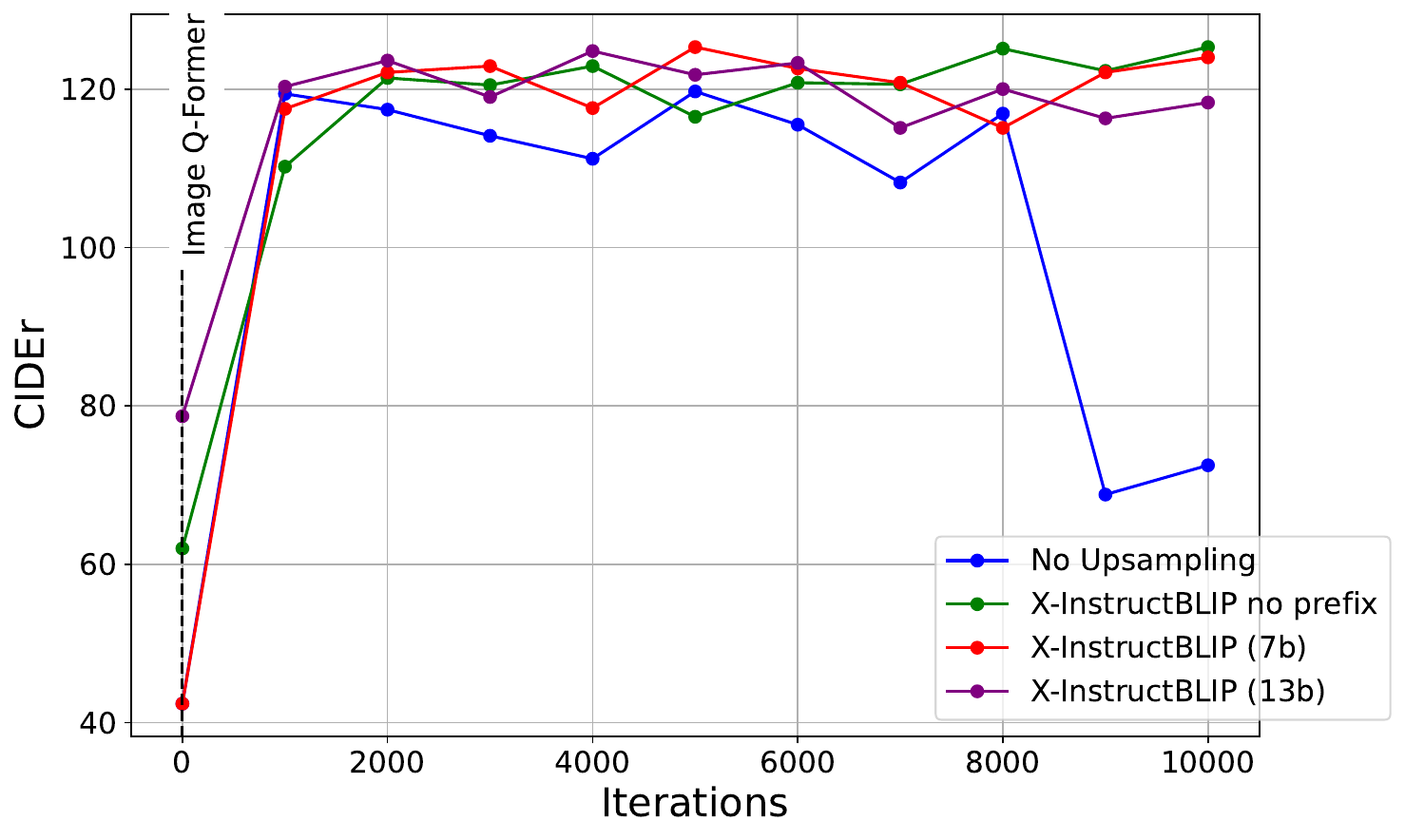}
    \caption{CIDEr score on MSVD (out-domain) over training iterations on Video Q-Former initialized from Image Q-Former. Most performance gains are achieved within only 1000 iterations.}
    \label{fig:iterations_msvd}
\end{figure}

Similarly, the Vicuna13b video Q-Former is initialized from the best checkpoint of the Vicuna13b Image Q-Former and validated every 1k iterations. While we let the Vicuna7b and 13b video Q-Formers train for 15k and 25k respectively, we observe early convergence at 15k and 5k iterations likely due to the pre-initialization with the Image Q-Former. During training, 5 frames are sampled for the Vicuna7b Video Q-Former, while 4 frames are sampled for the Vicuna13b to reduce computational demands. Figure \ref{fig:iterations_msvd} shows that the video performance converges in 1k iterations on an out of domain video captioning dataset. 

The best training approach for each model was empirically identified, and it is beyond the scope of the paper to rigorously analyze the reasons of the differences in training across modalities. We leave this to future work. 

\begin{table}[h]
\fontsize{5pt}{5pt}\selectfont
    \centering
 
    \begin{tabular}{ccccc}
    \toprule
    (7b/13b) &Image & Audio & 3D & Video$^\ast$\\\midrule
    Iterations & 775k/880k & 65k/300k & 65k/300k & 15k/5k \\
    Batch Size & 64/16 & 64/16 & 128/32 & 32/8 \\
      \bottomrule
    \end{tabular}
    \caption{Training hyperparameters. $^\ast$ Video projection is initialized from Image Projection. Parameters for 7b/13b model respectively.}
    \label{tab:implementation_param}
\end{table}

\begin{table}[tb]
\fontsize{7pt}{8pt}\selectfont
\centering

   \begin{tabular}{llll}
\toprule
        & MSVD  & VATEX  & MSVD QA  \\
        
        & {\tiny\textit{test}}  & {\tiny\textit{val} } & \textit{test} \\\noalign{\hrule height .3pt}
        
  \rowcolor{gray!10} X-Instruct Proj. (7b) & 118.2 & 58.5& 52.5  \\ \noalign{\hrule height .3pt}
  
  \rowcolor{gray!10} X-Instruct Proj. (7b)~{-upsample} & 73.3\verytiny{($\downarrow$44.9)}&41.6\verytiny{($\downarrow$16.9)} & 49.1\verytiny{($\downarrow$3.4)} \\
\bottomrule
    \end{tabular}
    \caption{Effect of MSRVTT Upsampling (at 10k iterations)}
 \label{tab:upsample_msrvtt}
\end{table}

\begin{table}[tb]
\fontsize{5pt}{5pt}\selectfont
\centering

   \begin{tabular}{llllllll}
\toprule
& \multicolumn{4}{c}{Zero-Shot} & \multicolumn{3}{c}{In-Domain}\\
& Flickr30k & NoCaps & VizWiz & GQA& OKVQA & \multicolumn{2}{c}{COCO} \\
& {\tiny\textit{test}} & {\tiny\textit{val-all}} & {\tiny\textit{test-dev}} & {\tiny\textit{balanced test-dev}}  & {\tiny\textit{test}} & {\tiny\textit{val}} & {\tiny\textit{test}}\\
  \rowcolor{gray!10}X-Instruct Proj.~(7b) & 82.1  & 117.7 & 34.9 & 48.1  & 52.5 &137.7 & 138.2 \\
  \rowcolor{gray!10}X-Instruct Proj.~(7b)-coco &79.4\verytiny{($\downarrow$2.7)}  & 116.5\verytiny{($\downarrow$1.2)} & 34.2\verytiny{($\downarrow$0.7)}& 48.2\verytiny{($\uparrow$0.1)}&  52.3\verytiny{($\downarrow$0.2)} &133.5\verytiny{($\downarrow$4.2)} &134.3\verytiny{($\downarrow$3.9)} \\
  \rowcolor{gray!10}X-Instruct Proj.~(13b) & 74.7 &  114.5 & 36.0 & 49.2  & 51.9 & 128.2 & 128.7\\
  \rowcolor{gray!10}X-Instruct Proj.~(13b)+coco &83.8\verytiny{($\uparrow$9.1)} &118.7\verytiny{($\uparrow$4.2)}&32.0\verytiny{($\downarrow$4.0)} &47.0\verytiny{($\downarrow$2.2)} & 44.6\verytiny{($\downarrow$7.3)}& 138.2\verytiny{($\uparrow$10.0)}&139.0\verytiny{($\uparrow$10.3)}\\

\bottomrule
    \end{tabular}
    \caption{Effect of COCO upsampling.} 
 \label{tab:coco_uspample}
\end{table}

\section{Evaluation Hyperparameters}
\label{app:evaluation_hyperparameters}

During the evaluation of X-InstructBLIP, we adhere to a consistent set of hyperparameters, with minor variations to accommodate the distinct needs of each task. A comprehensive list of these configurations is presented in table \ref{table:evaluation_hyperparameters}. In every experiment, we utilize Beam Search for generation, setting the beam size to 5, repetition penalty and temperature equal to 1.5 and 1 respectively. For tasks involving contrastive reasoning across video-audio modalities, a balanced representation and computational efficiency are achieved by querying two frames from both video and audio. The length penalty is typically configured to 1 for long caption tasks, -1 for Visual Question Answering (VQA) tasks requiring short answers, and 0 for short caption tasks. The minimum and maximum length constraints are adapted based on the task: for captions, we maintain a range of 10 to 80; for short-answer VQA tasks, the range is set from 1 to 10; for variable-length captions, the range is between 1 and 80. In the case of the InstructBLIP baseline for video datasets, we borrow the recommended inference setup of sampling 4 frames for the captioning baselines of MSVD and VATEX with the prompt \texttt{\footnotesize A video that shows} and the same generation hyperparameters as X-InstructBLIP.

\begin{table}[tb]
 \fontsize{6pt}{6pt}\selectfont
\centering
\begin{tabular}{lp{2cm}p{2cm}p{3.2cm}p{.9cm}p{.5cm}p{.5cm}}
\toprule
 & \textbf{Dataset}  & \textbf{Split} & \textbf{Prompt} & \textbf{Len. Penalty} & \textbf{Min Len.}& \textbf{Max Len.} \\\midrule
 \multirow{8}{*}{\rotatebox[]{90}{Image}} & Flickr30k~\cite{van2007flickr}  &  \multicolumn{1}{l}{\begin{tabular}{l}
      \textit{test}: 1,000 images
 \end{tabular}}
 & A short description & 1.  & 10  & 80 \\\cmidrule{2-7}
 & NoCaps~\cite{agrawal2019nocaps}  &  \multicolumn{1}{l}{\begin{tabular}{l}
 \textit{val}: 4,500 images \\
\textit{out-domain}: 1,413 images
  \end{tabular}}
 & A short description & 1.  & 10  & 80 \\\cmidrule{2-7}
& \underline{COCO} $^{\ast}$\cite{changpinyo2021conceptual} & \multicolumn{1}{l}{\begin{tabular}{l}
\textit{train}: 566,747 image-caption pairs \\
 \textit{val}: 5,000 images \\
  \textit{test}: 5,000 images \\
  \end{tabular}}& A short description. & 1.  & 10  & 80 \\\cmidrule{2-7}
   & VizWiz~\cite{bigham2010vizwiz} &   \multicolumn{1}{l}{\begin{tabular}{l}
\textit{test-dev}: 8,000 image-question pairs
  \end{tabular}} 
   & based on the given image respond to \{question\}  & -1.  & 1 & 10  \\\cmidrule{2-7}
  & \underline{OKVQA}~\cite{okvqa}  &\multicolumn{1}{l}{\begin{tabular}{l}\textit{test}: 5,046 examples \\ \end{tabular}}&  based on the given image respond to \{question\} answer & -1.  & 1 & 10 \\\cmidrule{2-7}
    & GQA~\cite{hudson2019gqa}  &  \multicolumn{1}{l}{\begin{tabular}{l}
\textit{balanced test-dev}: 12,578\\image-question pairs
  \end{tabular}} 
  & based on the given image respond to \{question\} & -1.  & 1 & 10  \\\cmidrule{2-7}
\multirow{2}{*}{\rotatebox[]{90}{3D} } & Modelnet40~\cite{wu20153d}  &
\multirow{2}{*}{\begin{tabular}{l}
\textit{test}: 2,468 point clouds
  \end{tabular}}
  & Describe the 3d model. \textcolor{blue}{A 3d model of} & -1. & 1 & 3\\\cmidrule(r{1pt}){2-2} \cmidrule(l{1pt}){4-7}
& Modelnet40\textdagger && Describe the 3d model. & 0. & 10 & 80\\\midrule

\multirow{2}{*}{\rotatebox[]{90}{Audio}} & Clotho~\cite{drossos2020clotho}   &\multicolumn{1}{l}{\begin{tabular}{l}
 \textit{eval} (v1): 1,045 audios \\
\textit{val} (v2): 1,045 audios
  \end{tabular}}& A short description. & 0.  & 10 & 80\\\cmidrule{2-7}
    &ClothoAQA~\cite{lipping2022clotho}   &\multicolumn{1}{l}{\begin{tabular}{l}
 \textit{test}: 2,838\\ audio-question pairs \\
  \end{tabular}}& \{question\} & -1.  & 1 & 10\\\cmidrule{2-7}
  &ESC50~\cite{piczak2015esc}   &\multicolumn{1}{l}{\begin{tabular}{l}
 \textit{test}: 2,000 audios \\
  \end{tabular}}& Describe the audio. \textcolor{blue}{An audio of} & 1.  & 10 & 80\\\cmidrule{2-7}
 & \underline{AudioCaps}$^\ast$~\cite{kim2019audiocaps}  & \begin{tabular}{l}
 \textit{train}: 38,695 audio-caption pairs \\ 
 \textit{val}: 380 audios
  \end{tabular}& A short description & 0. & 1 & 80\\\midrule

 \multirow{5}{*}{\rotatebox[]{90}{Video}} & MSVD~\cite{chen2011collecting}  &\begin{tabular}{l}
 \textit{test}: 670 images\footnotemark\\
  \end{tabular}& A short description &  1. & 10 & 80 \\\cmidrule{2-7}
 & \underline{MSRVTT}~$^\ast$\cite{xu2016msr} &\begin{tabular}{l}
     \textit{train}: 130,260 video-caption pairs \\
     \textit{val}:  497 videos\\
 \textit{test}: 2,990 videos
  \end{tabular}& A short description & 1. & 10 & 80\\\cmidrule{2-7}
 & MSVD QA~\cite{xu2017video} &\begin{tabular}{l}
 \textit{test}: 13,157 video-question pairs
  \end{tabular}
  &based on the given video respond to \{question\} &  -1. &1  & 10 \\\midrule

\multirow{2}{*}{\rotatebox[]{90}{A+V}} & MusicAVQA~\cite{li2022learning}  &\begin{tabular}{l}
 \textit{val}: 3,698 examples \\
  \textit{test}: 7,402 video-question pairs \\
  \end{tabular}& Question: \{question\} Answer: & -1. & 1 & 10\\\cmidrule{2-7}
    & {VATEX}\cite{wang2020vatex} & \begin{tabular}{l}
    \textit{val}:  3,000 videos\\

  \end{tabular} & A short description & 1. & 10 & 80\\

\bottomrule
 
\end{tabular}
\caption{Hyperparameters used on each of the evaluation datasets. \underline{Underlined} datasets are in-domain evaluations.
$^{\ast}$ datasets are used for best checkpoint selection. \textcolor{blue}{Blue} text is provided as input to the LLM but not the Q-Former.}
\label{table:evaluation_hyperparameters}
\end{table}

\section{Instruction Tuning Suite}
\label{app:instr_tuning_suite}

Table \ref{table:datasets_suite} presents a comprehensive list of datasets employed in the instruction tuning process for X-InstructBLIP, accompanied by their corresponding dataset sizes. Datasets labeled with $^{\ast\ast}$ have been generated automatically through the round-trip-consistency procedure. 
Datasets marked with $^\text{\textbullet}$ indicate instances of data loss resulting from file corruption or expired links.

\begin{table}[h]
 \fontsize{5pt}{5pt}\selectfont
\centering
\begin{tabular}{llll}
\toprule
 & \textbf{Task} & \textbf{Dataset}  & \textbf{Training Size} \\
\midrule
\multirow{ 12}{*}{\rotatebox[]{90}{Image} } &\multirow{ 6}{*}{Caption} & CapFilt14M~\cite{li2023blip}  & 13,873,136 image-caption pairs \\
\cmidrule{3-4}
& & Conceptual Captions 12M~\cite{changpinyo2021conceptual} & 6,029,862 image-caption pairs$^\text{\textbullet}$ \\
\cmidrule{3-4}
& & MS COCO Dataset~\cite{changpinyo2021conceptual}  & 566,747  image-caption pairs \\
\cmidrule{3-4}
& & SBU Captions~\cite{NIPS2011_5dd9db5e} & 859,739  image-caption pairs \\\cmidrule{3-4}
& & Visual Genome Captions~\cite{krishna2017visual}  & 821,774  image-caption pairs \\ \cmidrule{2-4}
& \multirow{ 5}{*}{QA} & AOK VQA~\cite{AOKVQA} & 17,056 question-answer pairs \\\cmidrule{3-4}
& & OK VQA~\cite{okvqa}  & 9,009  question-answer pairs \\
\cmidrule{3-4}
&& OCR VQA~\cite{mishraICDAR19} & 1,002,146  question-answer pairs \\
\cmidrule{3-4}
& &  Visual Genome QA~\cite{krishna2017visual}  & 1,440,069  question-answer pairs \\ \cmidrule{3-4}
& & VQAV2~\cite{balanced_vqa_v2}  & 658,104  question-answer pairs \\\cmidrule{2-4}
& Dialogue & LLaVA150k~\cite{liu2023visual} & 394,276 image-instruction pairs \\
\midrule\midrule

\multirow{4}{*}{\rotatebox[]{90}{Audio} } & \multirow{ 2}{*}{Caption} & AudioCaps~\cite{kim2019audiocaps}  & 38,701 audio-caption pairs$^\text{\textbullet}$ \\
\cmidrule{3-4}
&&WAVCaps~\cite{wavcaps} &  297,341 audio-caption pairs$^\text{\textbullet}$ \\ 
\cmidrule{2-4}
&QA & AudioCaps QA$^{\ast\ast}$  &  24,158 question-answer pairs \\\cmidrule{2-4}
& Classification & AudioSet {\tiny balanced train}~\cite{gemmeke2017audio} & 14,141 labeled audios$^\text{\textbullet}$ \\
\midrule\midrule

\multirow{2}{*}{\rotatebox[]{90}{3D} } & Caption & Cap3D~\cite{luo2023scalable}  & 651,576 point cloud-caption pairs\\
\cmidrule{2-4}
&QA & Cap3D QA$^{\ast\ast}$ &  250,070 question-answer pairs \\\midrule\midrule

\multirow{4}{*}{\rotatebox[]{90}{Video} } & \multirow{ 2}{*}{Caption} & MSRVTT~\cite{xu2016msr}  & 130,260 video-caption pairs \\
\cmidrule{3-4}
&&WebVid2M~\cite{bain2021frozen} & 2M video-caption pairs \\ 
\cmidrule{2-4}
&QA & MSRVTT QA~\cite{xu2017video} &  149,075 question-answer \\\bottomrule
\end{tabular}
\caption{Datasets for Instruction Tuning: This table presents datasets used for instruction tuning, along with their associated task types and sizes. $^\text{\textbullet}$Missing data results from expired links and corrupted files. $^{\ast\ast}$ Datasets marked with double asterisks are generated automatically within this study.}
\label{table:datasets_suite}
\end{table}

\section{Prompt Templates}
\label{app:templates}

X-InstructBLIP has undergone fine-tuning using a diverse array of instruction templates,  tailored to cover a wide spectrum of tasks and modalities. For reference, the specific templates corresponding to each modality can be found in the following tables: Table \ref{table:instr_tuning_templates_image} for images, Table \ref{table:instr_tuning_templates_audio} for audio, Table \ref{table:instr_tuning_templates_3d} for 3D, and Table \ref{table:instr_tuning_templates_video} for videos. Compared to InstructBLIP~\cite{dai2023instructblip} caption templates have increased from 13 to 32, while question-answering templates have grown from 10 to 21. These enhancements have been strategically incorporated to foster greater adaptability of the model to a wide range of user instructions.

\begin{table}[h]
\centering
\fontsize{5}{5}\selectfont
\begin{tabular}{l|l}
\toprule
\multicolumn{2}{c}{\textbf{Image Instruction Templates} } \\
\midrule
\rotatebox[]{90}{QA} & 
\begin{tabular}[l]{@{}l@{}}
 ``\{question\}" \\
  ``Q: \{question\} A:''\\
  ``Answer the following question: \{question\}''\\
  ``Question: \{question\} Answer:''\\
  ``How would you answer \{question\}?''\\
  ``What is the answer to the question \{question\}?''\\
  ``Answer the question based on the image. Question: \{question\} Answer: ''\\
  ``Instruction: Answer the following question by reference to the input image. Question: \{question\} Answer:''\\
  ``Given the photo, what is the answer to the question \{question\}?''\\
  ``What's your response to the query \{question\}?''\\
  ``Please provide an answer to \{question\}''\\
  ``Respond to the query \{question\}''\\
    ``Based on the given image, respond to \{question\}''\\
  ``Question: \{question\} What's your response?''\\
  ``Consider the following query: \{question\}''\\
  ``Could you help answer the question \{question\}?''\\
    ``Referencing the provided image, can you answer the question \{question\}?''\\
    ``With respect to the image shown, please answer \{question\}''\\
    ``What's your answer to \{question\} in the context of the provided image?''\\
  ``Question (refer to the image for context): \{question\} Answer:''\\
    ``In response to the question \{question\}, what would your answer be?"
\end{tabular} \\
\midrule
\rotatebox[]{90}{Caption} & 
\begin{tabular}[l]{@{}l@{}}
 ``A short caption:''\\
 ``A short description:''\\
 ``A photo of''\\
 ``A photo that shows''\\
 ``A picture of''\\
 ``A picture that shows''\\
 ``An image of''\\
 ``A image that shows''\\
 ``Write a short description.''\\
 ``Write a description for the image.''\\
 ``Provide a description of what is presented in the image.''\\
 ``Briefly describe the content of the image.''\\
 ``Can you briefly explain what you see in the image?''\\
 ``Could you use a few words to describe what you perceive in the image?''\\
 ``Please provide a short description of the image.''\\
 ``Using language, provide a short account of the image.''\\
 ``Use a few words to illustrate what is happening in the photo."
 ``Write a description for the photo.''\\
 ``Provide a description of what is presented in the photo.''\\
 ``Briefly describe the content of the photo.''\\
 ``Can you briefly explain what you see in the photo?''\\
 ``Could you use a few words to describe what you perceive in the photo?''\\
 ``Please provide a short description of the picture.''\\
 ``Using language, provide a short account of the picture.''\\
 ``Use a few words to illustrate what is happening in the picture.''\\
 ``Write a description for the picture.''\\
 ``Provide a description of what is presented in the picture.''\\
 ``Briefly describe the content of the picture.''\\
 ``Can you briefly explain what you see in the picture?''\\
 ``Could you use a few words to describe what you perceive in the picture?''\\
 ``Please provide a short description of the picture.''\\
 ``Using language, provide a short account of the picture.''\\
 ``Use a few words to illustrate what is happening in the picture."
\end{tabular} \\\bottomrule
\end{tabular}
\caption{Instruction-tuning templates for image tasks}
\label{table:instr_tuning_templates_image}
\end{table}

\begin{table}[h]
\centering
\fontsize{5}{5}\selectfont
\begin{tabular}{l|l}
\toprule
\multicolumn{2}{c}{\textbf{Audio Instruction Templates} } \\
\midrule
\rotatebox[]{90}{QA} & 
\begin{tabular}[l]{@{}l@{}}
``\{question\}\\
``Question: \{question\} Answer:''\\
``Q: \{question\} A:''\\
``Based on the audio, \{question\}''\\
``Answer the following question based on the audio: \{question\}''\\
``Question: \{question\} Provide an answer based on the audio.''\\
``How would you answer \{question\} based on the audio?''\\
``What is the answer to the question \{question\} using the audio as a reference?''\\
``Answer the question using the audio. Question: \{question\} Answer: ''\\
``Instruction: Answer the following question by referencing the audio. Question: \{question\} Answer:''\\
``Given the audio, what is the answer to the question \{question\}?''\\
``What's your response to the query \{question\} considering the audio?''\\
``Please provide an answer to \{question\} using the audio as context.''\\
``Respond to the query \{question\} based on the audio content.''\\
``Based on the provided audio, respond to \{question\}''\\
``Question: \{question\} What's your response using the audio for context?''\\
``Consider the following query and the audio: \{question\}''\\
``Could you help answer the question \{question\} using the audio as reference?''\\
``Referencing the provided audio, can you answer the question \{question\}?''\\
``With respect to the audio provided, please answer \{question\}''\\
``What's your answer to \{question\} in the context of the provided audio?''\\
``Question (refer to the audio for context): \{question\} Answer:''\\
``In response to the question \{question\}, what would your answer be based on the audio?''\\
``Given the audio, how would you respond to \{question\}?''\\
``Taking the audio into consideration, what is your response to \{question\}?''\\
``Based on the audio, how would you answer \{question\}?"
\end{tabular} ''\\
\midrule
\rotatebox[]{90}{Classification} & 
\begin{tabular}[l]{@{}l@{}}
``Classify the following audio:''\\
``What is the category of this audio clip?''\\
``Identify the content of the following audio:''\\
``Provide a classification for the audio.''\\
``Analyze and categorize the following audio.''\\
``Describe the category of the given audio.''\\
``Determine the type of this audio clip.''\\
``Can you classify what you hear in the audio?''\\
``What type of audio is this?''\\
``How would you classify this audio clip?''\\
``Please identify the category of the following audio:''\\
``What category does the following audio fall into?''\\
 ``Classify the sounds in this audio clip."
\end{tabular} ''\\
\midrule
\rotatebox[]{90}{Caption} & 
\begin{tabular}[l]{@{}l@{}}
``A short caption:''\\
``A short description:''\\
``An audio of''\\
``An audio that shows''\\
``Write a short description.''\\
``Write a description for the audio.''\\
``Provide a description of what is presented in the audio.''\\
``Briefly describe the content of the audio.''\\
``Can you briefly explain what you hear in the audio?''\\
``Could you use a few words to describe what you perceive in the audio?''\\
``Please provide a short description of the audio.''\\
``Using language, provide a short account of the audio.''\\
``Use a few words to illustrate what is happening in the audio.''\\
``Describe briefly the contents of the audio.''\\
``Please provide a brief summary of the audio.''\\
``What does the audio contain?''\\
``What can you hear in the audio?''\\
``What sounds are present in the audio?''\\
``Summarize the audio in a few words.''\\
``Write a brief summary of the audio content.''\\
``Could you provide a concise explanation of the audio's contents?''\\
``Describe what the audio represents.''\\
``What is the audio depicting?''\\
``In a few words, describe what you hear in the audio."
\end{tabular} ''\\\bottomrule
\end{tabular}
\caption{Instruction-tuning templates for audio tasks}
\label{table:instr_tuning_templates_audio}
\end{table}

\begin{table}[h]
\centering
\fontsize{5}{5}\selectfont
\begin{tabular}{l|l}
\toprule
\multicolumn{2}{c}{\textbf{3D Instruction Templates }} \\
\midrule
\rotatebox[]{90}{QA} & 
\begin{tabular}[l]{@{}l@{}}
``\{question\}''\\
``Question: \{question\} Answer:''\\
``Q: \{question\} A:''\\
``Based on the 3D model, \{question\}''\\
``Answer the following question based on the 3D model: \{question\}''\\
``Question: \{question\} Provide an answer based on the 3D model.''\\
``How would you answer \{question\} based on the 3D model?''\\
``What is the answer to the question \{question\} using the 3D model as a reference?''\\
``Answer the question using the 3D model. Question: \{question\} Answer: ''\\
``Instruction: Answer the following question by referencing the 3D model. Question: \{question\} Answer:''\\
``Given the 3D model, what is the answer to the question \{question\}?''\\
``What's your response to the query \{question\} considering the 3D model?''\\
``Please provide an answer to \{question\} using the 3D model as context.''\\
``Respond to the query \{question\} based on the 3D model content.''\\
``Based on the provided 3D model, respond to \{question\}''\\
``Question: \{question\} What's your response using the 3D model for context?''\\
``Consider the following query and the 3D model: \{question\}''\\
``Could you help answer the question \{question\} using the 3D model as reference?''\\
``Referencing the provided 3D model, can you answer the question \{question\}?''\\
``With respect to the 3D model provided, please answer \{question\}''\\
``What's your answer to \{question\} in the context of the provided 3D model?''\\
``Question (refer to the 3D model for context): \{question\} Answer:''\\
``In response to the question \{question\}, what would your answer be based on the 3D model?''\\
``Given the 3D model, how would you respond to \{question\}?''\\
``Taking the 3D model into consideration, what is your response to \{question\}?''\\
``Based on the 3D model, how would you answer \{question\}?"
\end{tabular} \\
\midrule
\rotatebox[]{90}{Caption} & 
\begin{tabular}[l]{@{}l@{}}
``A short caption:''\\
``A short description:''\\
``A 3D model of''\\
``A 3D model that shows''\\
``Write a short description.''\\
``Write a description for the 3D model.''\\
``Provide a description of what is presented in the 3D model.''\\
``Briefly describe the content of the 3D model.''\\
``Can you briefly explain what you see in the 3D model?''\\
``Could you use a few words to describe what you perceive in the 3D model?''\\
``Please provide a short description of the 3D model.''\\
``Using language, provide a short account of the 3D model.''\\
``Use a few words to illustrate what is happening in the 3D model.''\\
``Describe briefly the contents of the 3D model.''\\
``Please provide a brief summary of the 3D model.''\\
``What does the 3D model contain?''\\
``What can you identify in the 3D model?''\\
``What structures are present in the 3D model?''\\
``Summarize the 3D model in a few words.''\\
``Write a brief summary of the 3D model content.''\\
``Could you provide a concise explanation of the 3D model's contents?''\\
``Describe what the 3D model represents.''\\
``What is the 3D model depicting?''\\
``In a few words, describe what you see in the 3D model."
\end{tabular} \\\bottomrule
\end{tabular}
\caption{Instruction-tuning templates for 3D tasks}
\label{table:instr_tuning_templates_3d}
\end{table}

\begin{table}[h]
\centering
\fontsize{5}{5}\selectfont
\begin{tabular}{l|l}
\toprule
\multicolumn{2}{c}{\textbf{Video Instruction Templates}  } \\
\midrule
\rotatebox[]{90}{QA} & 
\begin{tabular}[l]{@{}l@{}}
``Given the video, \{question\}''\\
``Q: \{question\} A:''\\
``Answer the following question based on the video: \{question\}''\\
``Question: \{question\} Answer:''\\
``How would you answer \{question\} after watching the video?''\\
``What is the answer to the question \{question\} after viewing the video?''\\
``Answer the question based on the video. Question: \{question\} Answer: ''\\
``Instruction: Answer the following question by reference to the input video. Question: \{question\} Answer:''\\
``Given the video, what is the answer to the question \{question\}?''\\
``What's your response to the query \{question\} after watching the video?''\\
``Please provide an answer to \{question\} after watching the video''\\
``Respond to the query \{question\} based on the video''\\
``Based on the given video, respond to \{question\}''\\
``Question: \{question\} What's your response after watching the video?''\\
``Consider the following query: \{question\}''\\
``Could you help answer the question \{question\}?''\\
``Referencing the provided video, can you answer the question \{question\}?''\\
``With respect to the video shown, please answer \{question\}''\\
``What's your answer to \{question\} in the context of the provided video?''\\
``Question (refer to the video for context): \{question\} Answer:''\\
``In response to the question \{question\}, what would your answer be after viewing the video?"
\end{tabular} \\
\midrule
\rotatebox[]{90}{Caption} & 
\begin{tabular}[l]{@{}l@{}}
``A short caption for the video:''\\
``A short description of the video:''\\
``A video of''\\
``A video that shows''\\
``Describe the video briefly.''\\
``Write a description for the video.''\\
``Provide a description of what is presented in the video.''\\
``Briefly describe the content of the video.''\\
``Can you briefly explain what you see in the video?''\\
``Could you use a few words to describe what you perceive in the video?''\\
``Please provide a short description of the video.''\\
``Using language, provide a short account of the video.''\\
``Use a few words to illustrate what is happening in the video."
\end{tabular} \\\bottomrule
\end{tabular}
\caption{Instruction-tuning templates for audio tasks}
\label{table:instr_tuning_templates_video}
\end{table}

\section{Ethics Statement}

In this research, we present a framework for aligning multiple modalities with a frozen large language model (LLM). Our methodology strictly involves the use of publicly available and free datasets, ensuring we do not engage in the collection of private data. However, it is crucial to acknowledge that publicly sourced datasets carry implicit biases~\cite{fabbrizzi2022survey,yeh2023evaluating,motoki2023more}. These biases reflect historical and societal inequalities, potentially influencing the model's outputs. Our framework builds upon a pre-existing frozen LLM. While this approach benefits from the extensive knowledge encoded within the LLM, it is important to recognize that such models can inherently propagate biases present in their training data. Additionally, there is a non-negligible risk of generating false or misleading information. While there exist tools to measure language model toxicity such as Helm~\cite{liang2023holistic}, their evaluation datasets are constrained in the language modality, and hence are not applicable to measure toxicity across modalities which is the focus of this work. We leave the generation of cross-modal datasets for toxicity and bias measurement as a future research direction.

Users of our framework should be aware of these limitations and exercise caution, particularly in applications where the accuracy and impartiality of outputs are critical. We advocate for responsible use of our framework, especially in sensitive contexts. Users should critically assess and verify the model's outputs and consider the potential for reinforcing biases or spreading misinformation. Furthermore, we commit to transparency regarding our model's capabilities and limitations. All code, data, and model weights will be released to ensure reproducibility and encourage external evaluation and subsequent research.

\section{Reproducibility Statement}
In alignment with the principles of open science and to foster reproducibility, transparency, and further research, we promise to provide open source access to all the resources associated with our study, including: a complete, documented, and public codebase with all the scripts, models, preprocessing, and evaluation code necessary to replicate the experiments. We will be further releasing the pretrained model weights along side the exact evaluation configs that generated the results cited in the paper. We show our commitment to reproducibility through an extensive supplementary section that highlights details of training and evaluation. Furthermore, all experiments were completed with prespecified random seeds that will also be made available in the experiment configuration files. Finally, we will release all datasets collected for this study for public download, as well as the code used to generate them. In addition to providing these resources, we pledge to maintain them and offer requisite support for any queries or clarifications related to the provided resources, contributing to a supportive and inclusive research environment.

\newpage


%
%
\bibliographystyle{splncs04}
\bibliography{main}
\end{document}